\documentclass{article}

\PassOptionsToPackage{numbers, compress}{natbib}

\usepackage[preprint]{neurips_2026}

\usepackage[utf8]{inputenc} 
\usepackage[T1]{fontenc}    
\usepackage{hyperref}       
\usepackage{url}            
\usepackage{booktabs}       
\usepackage{amsfonts}       
\usepackage{nicefrac}       
\usepackage{microtype}      
\usepackage{xcolor}         
\usepackage[ruled,vlined]{algorithm2e}
\usepackage{multirow}
\usepackage{makecell}
\usepackage{times}
\usepackage{latexsym}
\usepackage{microtype}
\usepackage{inconsolata}
\usepackage{graphicx}
\usepackage{amssymb}
\usepackage{amsmath}
\usepackage{enumitem}
\usepackage[table]{xcolor}
\usepackage{caption}
\usepackage[most]{tcolorbox}
\usepackage{pifont}
\usepackage{tabularx}

\title{OASES: Outcome-Aligned Search-Evaluation Co-Training for Agentic Search}

\author{
\textbf{Erhan Zhang}$^{1}$\thanks{Equal contribution.}, 
\textbf{Yiqun Chen}$^{1}$\footnotemark[1], 
\textbf{Zechun Niu}$^{1}$, 
\textbf{Wei Yang}$^{3}$, 
\textbf{Xiaochi Wei}$^{2}$, \\
\textbf{Yan Gao}$^{2}$, 
\textbf{Yi Wu}$^{2}$, 
\textbf{Yao Hu}$^{2}$, 
\textbf{Jiaxin Mao}$^{1}$\thanks{Corresponding author.} \\
$^{1}$ Renmin University of China 
$^{2}$ Xiaohongshu Inc. 
$^{3}$ University of Southern California \\
\texttt{erhanzhang@ruc.edu.cn}, 
\texttt{chenyiqun990321@ruc.edu.cn}, \\
\texttt{maojiaxin@gmail.com}
}

\begin{document}

\maketitle

\begin{abstract}
Agentic search enables language models to solve knowledge-intensive tasks by adaptively acquiring external evidence over multiple steps. Reinforcement learning with verifiable rewards (RLVR) has emerged as a widely adopted training paradigm for search agents, yet outcome-only rewards are sparse and provide limited credit assignment for intermediate search actions. Existing process-reward methods therefore seek to densify supervision through proxy signals, external evaluators, or likelihood-based information gain. However, proxy rewards can deviate from the final outcome objective, while fixed evaluators can become stale as the search policy evolves, leading to unreliable process supervision. To address these challenges, we propose \textbf{OASES}, an \textbf{O}utcome-\textbf{A}ligned \textbf{S}earch-\textbf{E}valuation \textbf{S}upervision framework for agentic search. OASES derives outcome-aligned process rewards by evaluating how well each intermediate search state supports answering the original question. It further co-trains the search policy and the state evaluator on policy, allowing the evaluator to adapt to evolving search behavior and provide more reliable process rewards. Experiments on five multi-hop QA benchmarks show that OASES consistently outperforms strong RL baselines, with further analyses confirming the benefits of outcome-aligned process rewards and search-evaluation co-training. 
\end{abstract}

\begin{figure*}[t]
    \centering
    \includegraphics[width=\linewidth]{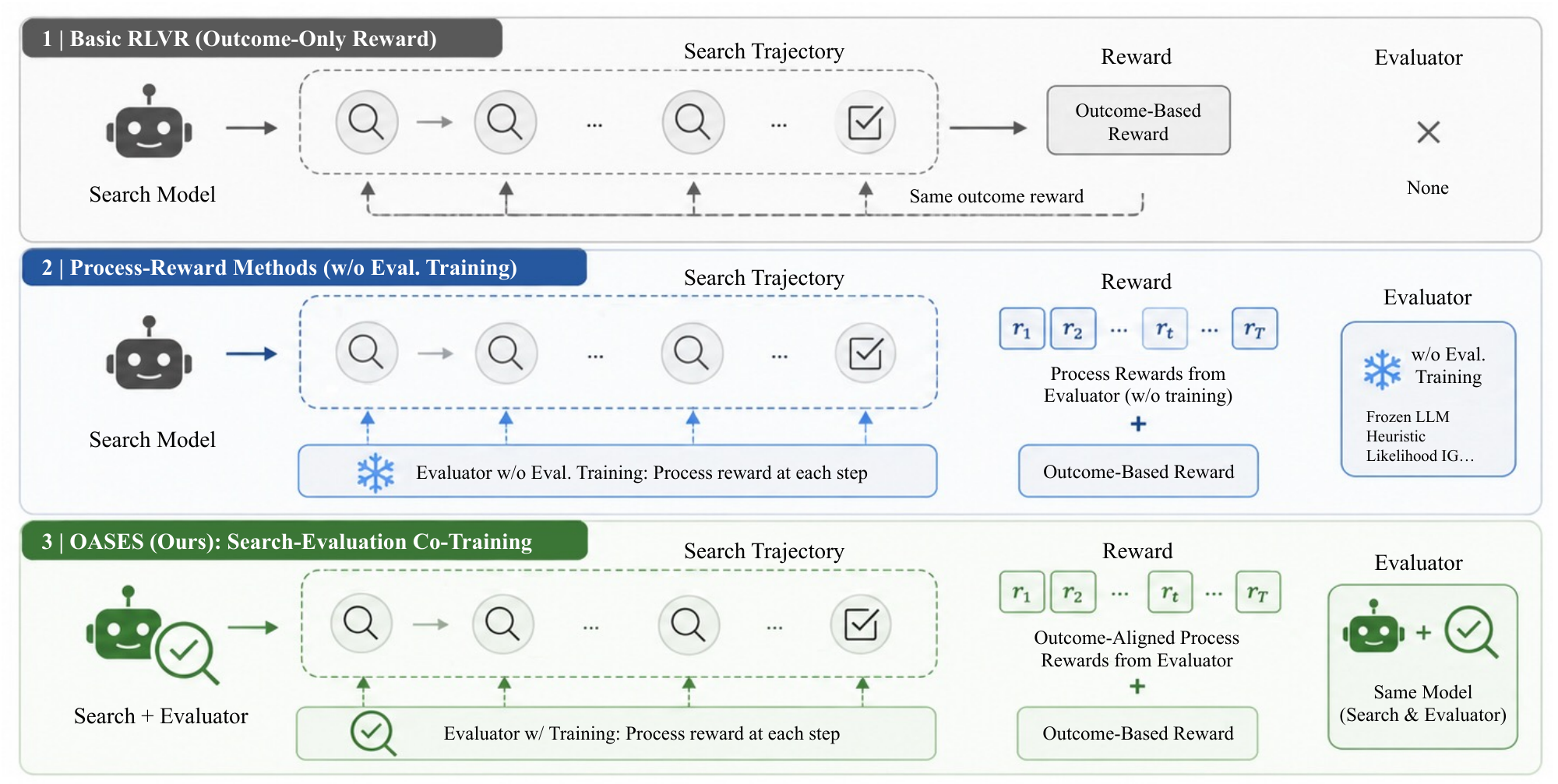}
    \caption{
    Comparison of reward designs for RLVR-based agentic search. Outcome-only RLVR provides trajectory-level feedback. Existing process-reward methods add step-level rewards from a separate evaluator without evaluation training. OASES co-trains a single policy for search and state evaluation, producing outcome-aligned process rewards while reducing evaluator--policy mismatch.
    }
    \label{fig:oases_framework}
\end{figure*}

\section{Introduction}

Large language models (LLMs) are increasingly moving beyond one-shot generation toward \emph{agentic} problem solving, where a model interacts with external tools and environments over multiple steps~\cite{yao2022react, schick2023toolformer}. Among these paradigms, \emph{agentic search} has become an important framework for knowledge-intensive question answering, allowing models to actively acquire task-relevant evidence~\cite{li2025search, jin2025search, song2025r1}. Unlike standard retrieval-augmented generation (RAG), which typically relies on a fixed retrieval stage~\cite{lewis2020retrieval}, agentic search adaptively acquires information needed for problems that cannot be solved from parametric memory alone.

Recent work has explored reinforcement learning (RL) to improve the search and reasoning abilities of such agents. In particular, reinforcement learning with verifiable rewards (RLVR)~\cite{lambert2024tulu, guo2025deepseek} provides a natural training paradigm, since many knowledge-intensive tasks admit automatically checkable outcome signals, such as exact match or F1. By optimizing the outcome objective, RLVR avoids reliance on supervised trajectories and allows the agent to discover its own search strategies. However, applying RLVR to agentic search remains challenging from both the \emph{search} and \emph{evaluation} sides.

On the search side, the key challenge is to teach the agent how to search by telling which intermediate actions are truly useful~\cite{lin2025comprehensive}. A search trajectory may contain multiple rounds of reasoning, query formulation, evidence acquisition, and answer refinement, but outcome rewards provide feedback only after the final answer is produced. Such sparse and delayed supervision makes it difficult to assign credit to individual search decisions~\cite{tan2025process, wang2026prorag, xuprinciple}. As a result, RLVR for agentic search naturally calls for dense process supervision that can guide the policy at intermediate search states.

On the evaluation side, however, reliable process supervision is difficult to obtain. Existing methods therefore rely on proxy signals, such as retrieved-document similarity~\cite{zheng2025stepsearch, petcu2026subsearch}, external LLM evaluators~\cite{xia2026search, zhao2025r, wang2026enhancing}, or heuristic rules~\cite{shi2025search}. While these signals provide denser feedback, they often optimize intermediate criteria such as relevance, format compliance, or judge preference, rather than whether the search state improves the final answer under the outcome objective. \emph{In short, proxy process supervision provides denser feedback, but it is not aligned with the outcome objective.}

Another group of methods defines process rewards through information gain, for example by measuring how the likelihood or perplexity of the gold answer changes after new information is acquired. Although these rewards are more directly tied to the final answer, their gain estimates can still be unreliable. Specifically, using fixed model to derive the likelihood and information gain may suffer from a distributional mismatch with the evolving search policy,~\cite{xie2026tips}, while directly using the policy model itself to score the intermediate search states can be inconsistent with the model's own generation behavior under partial search prefixes~\cite{wang2025information, liang2026ig, hu2026optimizing}. Optimizing such likelihood-based evaluation may reduce this mismatch, but risks answer leakage by fitting the gold answer before sufficient evidence is available. \emph{In short, likelihood-based information-gain rewards are closer to the outcome objective, but they can still provide inaccurate process supervision due to evaluator--policy mismatch.}

These limitations suggest that improving agentic search requires jointly addressing the search and evaluation sides. The search policy needs dense supervision to overcome sparse outcome feedback, while the evaluator must provide reliable feedback that remains aligned with the outcome objective and compatible with the evolving policy. This motivates a unified framework where state evaluation is not a separate scoring procedure, but a learnable capability optimized together with search.

In this paper, we propose \textbf{OASES}, an \textbf{O}utcome-\textbf{A}ligned \textbf{S}earch--\textbf{E}valuation \textbf{S}upervision framework for agentic search. OASES derives outcome-aligned process rewards by evaluating how well each intermediate search state supports answering the original question. Specifically, given the evidence accumulated up to a search state, an evaluator generates a state-conditioned answer, which is then scored by the same verifiable outcome metric used for the final answer. The improvement of this score across consecutive search states is used as a dense process reward, converting sparse outcome feedback into fine-grained credit assignment while keeping the reward signal aligned with the final objective. To reduce evaluator--policy mismatch, OASES further co-trains a shared-parameter model as both the search policy and the state evaluator. This allows the evaluator to adapt to the evolving search behavior and enables the policy to jointly improve its evidence acquisition and state-evaluation capabilities, without relying on a separate process reward model. Figure~\ref{fig:oases_framework} illustrates this distinction: compared with outcome-only RLVR and existing process-reward methods that rely on separate or non-optimized evaluators, OASES jointly optimizes search and state evaluation under a shared policy, yielding dense rewards that are both outcome-aligned and adaptive to the evolving search behavior.

We evaluate OASES on multi-hop QA benchmarks, showing that it consistently improves agentic search performance over strong RL baselines.
Further analyses demonstrate that outcome-aligned process rewards provide more effective credit assignment, and that search--evaluation co-training outperforms alternatives based on frozen evaluators or search-only optimization.

Our contributions are summarized as follows:
\begin{itemize}
\item We propose \textbf{OASES}, an outcome-aligned search--evaluation supervision framework that converts outcome improvements across search states into dense process rewards.
\item We introduce search--evaluation co-training, where a single policy learns both to search and to evaluate whether its accumulated information supports the final answer.
\item We conduct experiments on multi-hop QA benchmarks, showing that OASES improves agentic search over strong RL baselines and mitigates reward sparsity through more effective credit assignment.
\end{itemize}

\section{Related Work}

\paragraph{Agentic Search and RL-based Training.}

Recent advancements have focused on training LLMs as agentic search systems that interleave multi-step reasoning with external retrieval. Leveraging RL frameworks such as PPO~\cite{schulman2017proximal} or GRPO~\cite{shao2024deepseekmath}, representative agentic search methods such as Search-R1~\cite{jin2025search} and R1-Searcher~\cite{song2025r1} have demonstrated that such RL-based paradigms significantly bolster performance in complex search tasks. Parallel to these end-to-end agents, extensive research has investigated the optimization of specific search components, including planning~\cite{chen2025mao}, query reformulation~\cite{ma2023query, li2024dmqr, chen2025improving}, document selection~\cite{ke2024bridging, li2024rag}, and answer generation~\cite{yang2025qwen3}. More recently, there is a growing emphasis on the joint optimization of these modular components to achieve superior global performance~\cite{chen2026jade, chen2026beyond}. Orthogonal to training-based agentic search, RSE~\cite{wang2026not} studies how to recycle rollout-level experience at test time by reusing intermediate conclusions and failure patterns across rollouts.

\paragraph{Process Rewards for Agentic Search.}

Process supervision aims to alleviate the sparsity of outcome-only rewards by assigning feedback to intermediate search actions~\cite{lightman2023let, lin2025comprehensive}. Existing agentic-search methods obtain such feedback from several sources. Monte Carlo or tree-search methods propagate final outcome scores to intermediate decisions~\cite{zhang2025process}, but are often rollout-intensive and costly to integrate into online RL. Retrieval-oriented methods reward search steps using evidence relevance or coverage~\cite{petcu2026subsearch, shi2025search, zheng2025stepsearch}, while evaluator-based methods use learned reward models or external LLM judges to assess intermediate behavior~\cite{cui2025process, xuprinciple, zhao2025r, wang2026enhancing, xia2026search}. These approaches provide denser feedback, but they introduce separate scoring criteria or separate evaluators, which can be costly and may suffer from evaluator--policy mismatch. Closely related to our work, information-gain-based methods assign rewards according to whether a search state increases the likelihood of the gold answer~\cite{xie2026tips, wang2025information, liang2026ig, hu2026optimizing}. However, likelihood-based gains can be inaccurate, as they may be affected by evaluator--policy mismatch and fail to reflect whether the current state improves the model's final answer\footnote{We further discuss why these methods could not fully support a joint optimization of the search and evaluation models in Appendix~\ref{sec:joint_search_eval_discussion}.}.

\section{OASES: Outcome-Aligned Search-Evaluation Co-Training}
\begin{figure*}[t]
    \centering
    \includegraphics[width=\textwidth]{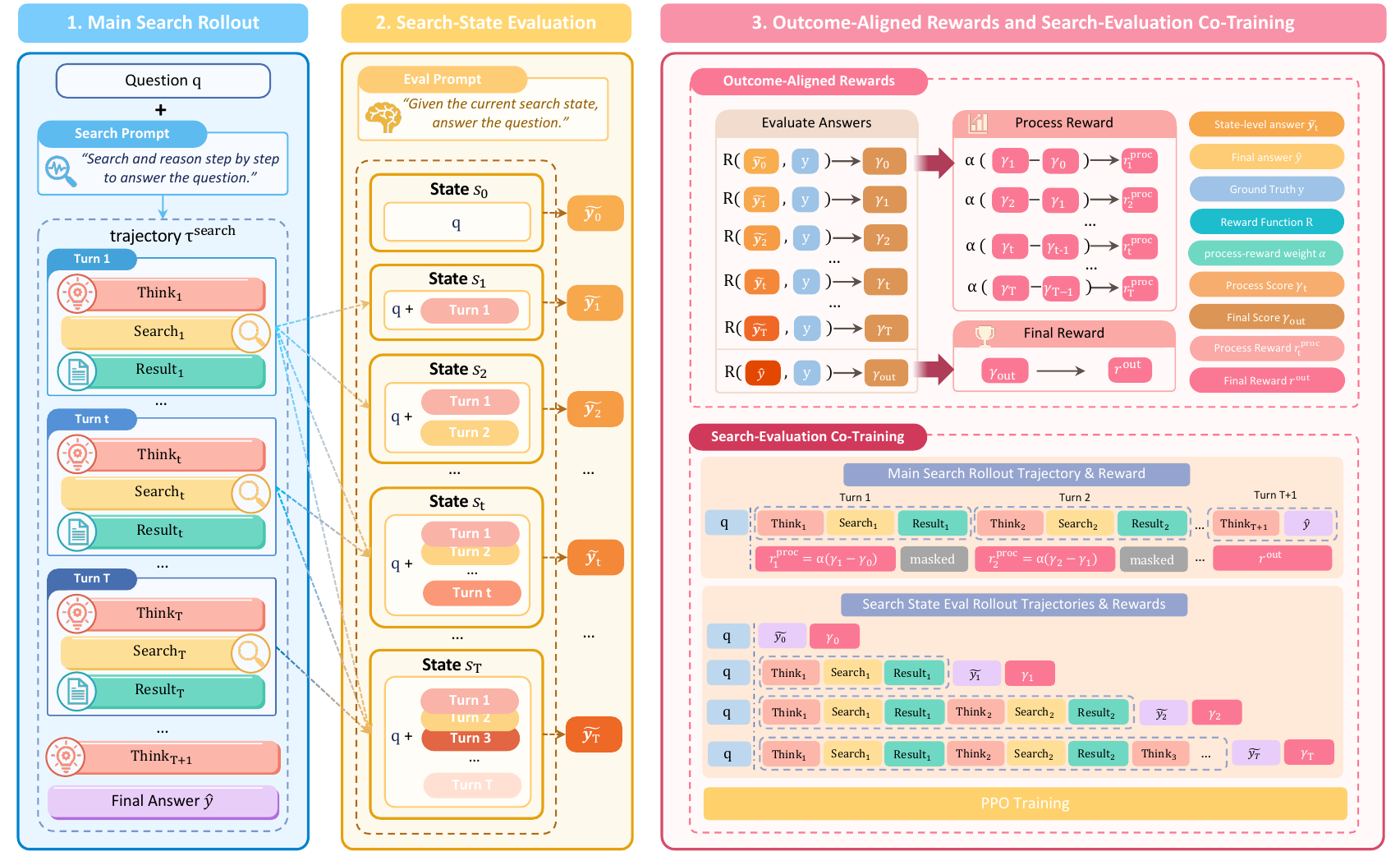}
    \caption{
    Overview of the OASES framework. The policy generates a main search trajectory $\tau^{\text{search}}$ and extracts an ordered sequence of search states $\{s_t\}_{t=0}^{T}$. For each state, the same policy generates a state-conditioned evaluation answer $\tilde{y}_t$, whose score $\gamma_t$ reflects how well the accumulated information supports the final answer. OASES derives process rewards from state-score differences $\gamma_t-\gamma_{t-1}$, combines them with the final outcome score $\gamma_{\mathrm{out}}$, and co-trains search and state evaluation with PPO.}
    \label{fig:framework}
\end{figure*}

This section presents OASES, an outcome-aligned search--evaluation framework for agentic search. As shown in Figure~\ref{fig:framework}, OASES uses the same verifiable outcome signal for state-level and final-answer evaluation, converting outcome improvements across search states into dense process rewards. It then co-trains a single policy for search and state evaluation with PPO~\cite{schulman2017proximal}.

\subsection{Preliminaries}

\paragraph{Task setup.}
Let $\mathcal{D}=\{(q_i, y_i)\}_{i=1}^{N}$ denote a training set of question--answer pairs, where $q_i$ is a question and $y_i$ is its ground-truth answer. For simplicity, we omit the sample index $i$ and write $(q,y)\sim\mathcal{D}$. The goal of agentic search is to learn a policy that, given a question $q$, interacts with an external search environment over multiple turns and produces a final answer that matches $y$.

\paragraph{Agentic search trajectory.}

We consider a multi-turn search setting with $T$ search turns. At each search turn $t$, the model generates a reasoning segment $c_t$ and a search action $a_t$, and then receives an observation $o_t$ from the environment. After the final search turn, the model generates a final reasoning segment $c_{T+1}$ and outputs an answer $\hat{y}$. The resulting main search trajectory is
\begin{equation}
\tau^{\text{search}}=\bigl(q,\{(c_t,a_t,o_t)\}_{t=1}^{T},c_{T+1},\hat{y}\bigr),
\label{eq:trajectory}
\end{equation}
where $c_t$, $a_t$, $o_t$, and $\hat{y}$ correspond to \texttt{<think>}, \texttt{<search>}, \texttt{<result>}, and \texttt{<answer>}, respectively, and $c_{T+1}$ is the final reasoning segment before answering. To expose the accumulated information available along the trajectory, we define a sequence of search states:
\begin{equation}
s_t=
\begin{cases}
\bigl(q,\{(c_{t'},a_{t'},o_{t'})\}_{t'=1}^{t}\bigr), & t=1,\dots,T,\\[3pt]
q, & t=0,
\end{cases}
\label{eq:search_state}
\end{equation}
where $s_t$ denotes the information state after the first $t$ search turns, and $s_0$ contains the original question without external search observation. Thus, $\{s_t\}_{t=0}^{T}$ forms an ordered sequence of search states, where each state contains exactly the information available to the model at that point.

\subsection{Search-State Evaluation}

OASES uses the same model $\pi_{\theta}$ for both the main search task and the search-state evaluation task, with different behaviors specified by different prompt templates. Let $P^{\text{search}}(q)$ denote the prompt for multi-turn agentic search, and let $P^{\text{eval}}(s_t)$ denote the prompt for search-state evaluation. Detailed prompt templates are provided in Appendix~\ref{app:prompt_templates}. In the evaluation prompt, the model answers the original question using only the information available in state $s_t$. For each question $q$, we construct the main search trajectory as
\begin{equation}
\tau^{\text{search}}
\sim
\pi_{\theta}(\cdot \mid P^{\text{search}}(q)),
\label{eq:main_rollout}
\end{equation}
where $\tau^{\text{search}}$ follows the trajectory format defined in Eq.~\eqref{eq:trajectory}. From this trajectory, we extract the ordered search states $\{s_t\}_{t=0}^{T}$ according to Eq.~\eqref{eq:search_state}. For each search state $s_t$, we construct a state-conditioned evaluation trajectory by prompting the same model to answer the original question, and collect all such trajectories as
\begin{equation}
\mathcal{T}^{\text{eval}}
=
\left\{
\tau_t^{\text{eval}}
=
\bigl(s_t,\tilde{y}_t\bigr)
\;:\;
\tilde{y}_t
\sim
\pi_{\theta}(\cdot \mid P^{\text{eval}}(s_t)),
\;
t=0,1,\dots,T
\right\}.
\label{eq:eval_trajectories}
\end{equation}
Here, $\tilde{y}_t$ is generated based only on the information contained in $s_t$. Together, $\tau^{\text{search}}$ and $\mathcal{T}^{\text{eval}}$ provide the trajectory-level information needed to estimate improvements across search states.

\subsection{Outcome-Aligned Process Rewards}

\paragraph{Answer quality scoring.}
Given the main search trajectory $\tau^{\text{search}}$ and the search-state evaluation trajectories $\mathcal{T}^{\text{eval}}$, OASES first scores all generated answers with the same verifiable answer scoring function $R(\cdot,\cdot)$. Specifically, we compute
\begin{equation}
\gamma_t = R(\tilde{y}_t, y),\quad t=0,1,\dots,T,
\qquad
\gamma_{\mathrm{out}} = R(\hat{y}, y),
\label{eq:answer_scores}
\end{equation}
where $\gamma_t$ is the state-level score of the evaluation answer $\tilde{y}_t$, and $\gamma_{\mathrm{out}}$ is the outcome score of the final answer $\hat{y}$. Since $\tilde{y}_t$ is generated using only the information in $s_t$, $\gamma_t$ measures how well the current search state supports answering the original question.

\paragraph{Rewards for the main search trajectory.}

As illustrated in Figure~\ref{fig:framework}, OASES converts the ordered state-level scores $\{\gamma_t\}_{t=0}^{T}$ into stepwise process rewards for the main search trajectory, and combines them with a final outcome reward. Intuitively, the $t$-th search turn is rewarded according to how much the newly accumulated information improves state answerability, while the final answer receives the standard outcome reward. Formally, we define
\begin{equation}
r_t^{\text{proc}}
=
\alpha\bigl(\gamma_t-\gamma_{t-1}\bigr),
\quad t=1,\dots,T,
\qquad
r^{\text{out}} = \gamma_{\mathrm{out}},
\label{eq:main_rewards}
\end{equation}
where $\alpha\ge 0$ controls the relative contribution of process rewards in the overall reward. We assign each process reward $r_t^{\text{proc}}$ to the end of the segment that triggers the $t$-th search interaction, and assign the outcome reward $r^{\text{out}}$ to the end of the final answer segment. Let $m_t$ and $m_{\text{ans}}$ denote these token positions, and let $L_{\text{search}}$ be the number of generated tokens in the main search rollout. The token-level reward sequence is
\begin{equation}
\rho_k^{\text{search}}
=
\sum_{t=1}^{T}
\mathbb{I}\!\left[k=m_t\right] r_t^{\text{proc}}
+
\mathbb{I}\!\left[k=m_{\text{ans}}\right] r^{\text{out}},
\qquad
k=1,\dots,L_{\text{search}}.
\label{eq:search_reward_assignment}
\end{equation}
As shown by the gray blocks in the ``Main Search Rollout Trajectory \& Reward'' row of Figure~\ref{fig:framework}, the PPO loss is applied only to model-generated tokens. We therefore mask out tokens from external search observations, i.e., the \texttt{<result>} segments, during optimization.

\paragraph{Rewards for search-state evaluation trajectories.}

In addition to supervising the main search rollout, OASES trains the model on the search-state evaluation trajectories.  For each $\tau_t^{\text{eval}}=(s_t,\tilde{y}_t)$, we use its state-level answer score $\gamma_t$ as the terminal reward. Let $m_t^{\text{eval}}$ denote the token position at the end of the evaluation answer $\tilde{y}_t$, and let $L_{\text{eval},t}$ be the number of generated tokens in the $t$-th evaluation rollout. The token-level reward sequence for this evaluation rollout is
\begin{equation}
\rho_k^{\text{eval},t}
=
\mathbb{I}\!\left[k=m_t^{\text{eval}}\right] \gamma_t,
\qquad
k=1,\dots,L_{\text{eval},t},
\quad
t=0,1,\dots,T.
\label{eq:eval_reward_assignment}
\end{equation}

Together, the search and evaluation objectives provide complementary supervision. The search objective rewards actions that improve state answerability, while the evaluation objective trains the model to judge whether the accumulated information supports the final answer. Because the process rewards are defined by score differences $\gamma_t-\gamma_{t-1}$, they telescope over the search trajectory and form an outcome-consistent redistribution of answerability improvement. We provide further analysis in Appendix~\ref{app:theory}.

\begin{algorithm*}[t]
\caption{OASES Training within PPO}
\label{alg:oases}
\small

\KwIn{training set $\mathcal{D}$, batch size $B$, policy $\pi_\theta$, value function $V_\phi$, process-reward weight $\alpha$}
\KwOut{updated policy parameters $\theta$ and value-function parameters $\phi$}

Initialize policy parameters $\theta$, critic parameters $\phi$, search buffer $\mathcal{B}^{\text{search}}$ and evaluation buffer $\mathcal{B}^{\text{eval}}$\;

\For{each training epoch}{
    \For{each batch $\mathcal{Q}\subset\mathcal{D}$ with $|\mathcal{Q}|=B$}{
        Clear search buffer $\mathcal{B}^{\text{search}}$ and evaluation buffer $\mathcal{B}^{\text{eval}}$\;
        \For{each question--answer pair $(q,y)\in\mathcal{Q}$}{
            Sample the main search trajectory $\tau^{\text{search}}$ by Eq.~\eqref{eq:main_rollout}, structured as Eq.~\eqref{eq:trajectory}\;
            
            Extract the search states $\{s_t\}_{t=0}^{T}$ according to Eq.~\eqref{eq:search_state}\;
            
            Sample the search-state evaluation trajectories $\mathcal{T}^{\text{eval}}=\{\tau_t^{\text{eval}}\}_{t=0}^{T}$ by Eq.~\eqref{eq:eval_trajectories}\;

            Compute state-evaluation scores $\{\gamma_t\}_{t=0}^{T}$ and final outcome score $\gamma_{\mathrm{out}}$ by Eq.~\eqref{eq:answer_scores}\;
            
            Compute the process rewards $\{r_t^{\text{proc}}\}_{t=1}^{T}$ and the outcome reward $r^{\text{out}}$ according to Eq.~\eqref{eq:main_rewards}\;

            Construct the token-level rewards $\rho^{\text{search}}$ by Eq.~\eqref{eq:search_reward_assignment}, and append $(\tau^{\text{search}},\rho^{\text{search}})$ to $\mathcal{B}^{\text{search}}$\;
            
            \For{$t=0$ \KwTo $T$}{
                Construct the token-level rewards $\rho^{\text{eval},t}$ by Eq.~\eqref{eq:eval_reward_assignment}, and append $(\tau_t^{\text{eval}},\rho^{\text{eval},t})$ to $\mathcal{B}^{\text{eval}}$\;
            }
        }
        Combine $\mathcal{B}^{\text{search}}$ and $\mathcal{B}^{\text{eval}}$ into a joint training buffer $\mathcal{B}$\;
        
       Compute returns and advantages from $\mathcal{B}$, then update $\pi_\theta$ and $V_\phi$ using Eq.~\eqref{eq:ppo_actor_loss} and Eq.~\eqref{eq:ppo_critic_loss}\;
    }
}
\end{algorithm*}

\subsection{Search-Evaluation Co-Training}

Algorithm~\ref{alg:oases} summarizes the PPO training procedure of OASES. A central design is to co-train search and evaluation with a single policy model $\pi_{\theta}$. For each training question, the model first acts as the search policy and generates a main search trajectory $\tau^{\mathrm{search}}$ with reward sequence $\rho^{\mathrm{search}}$.
This reward sequence trains the model to improve the final-answer outcome, while state-level process rewards assign outcome-aligned credit to intermediate search decisions.
We store these search rollouts in a search buffer $\mathcal{B}^{\mathrm{search}}$.

OASES then reuses the intermediate states along $\tau^{\mathrm{search}}$ to construct one state-conditioned evaluation rollout $\tau^{\mathrm{eval}}_t$ for each search state $s_t$, yielding $\mathcal{T}^{\mathrm{eval}}=\{\tau^{\mathrm{eval}}_t\}_{t=0}^{T}$ with reward sequences $\{\rho^{\mathrm{eval},t}\}_{t=0}^{T}$.
In these rollouts, the same model answers the original question conditioned on each state, and its reward is computed against the golden answer.
Thus, the evaluation rollouts train the model to produce better outcome-aligned answers from the information available at each state.
We store these evaluation rollouts in an evaluation buffer $\mathcal{B}^{\mathrm{eval}}$.

We compute return targets $\hat{G}$ and advantage estimates $\hat{A}$ over the joint buffer $\mathcal{B}=\mathcal{B}^{\mathrm{search}}\cup\mathcal{B}^{\mathrm{eval}}$ using generalized advantage estimation (GAE).
We also apply a format penalty to trajectories that violate the required interaction format, encouraging the model to follow the search protocol.
Let $r(\theta)=\pi_{\theta}(x\mid x_{<})/\pi_{\theta_{\mathrm{old}}}(x\mid x_{<})$ be the PPO importance ratio for a generated token $x$ conditioned on its prefix $x_{<}$.
The actor objective is
\begin{equation}
\mathcal{L}_{\mathrm{actor}}(\theta)
=
-\mathbb{E}_{(x,\hat{A})\sim \textcolor{red}{\mathcal{B}^{\mathrm{search}}\cup\mathcal{B}^{\mathrm{eval}}}}
\left[
\min\!\left(
r(\theta)\hat{A},\;
\mathrm{clip}\!\left(r(\theta),1-\epsilon,1+\epsilon\right)\hat{A}
\right)
\right],
\label{eq:ppo_actor_loss}
\end{equation}
where $\epsilon$ is the clipping coefficient.
The critic regresses return targets over the same joint buffer:
\begin{equation}
\mathcal{L}_{\mathrm{critic}}(\phi)
=
\mathbb{E}_{(x_{<},\hat{G})\sim \textcolor{red}{\mathcal{B}^{\mathrm{search}}\cup\mathcal{B}^{\mathrm{eval}}}}
\left[
\left(
V_{\phi}(x_{<})-\hat{G}
\right)^2
\right].
\label{eq:ppo_critic_loss}
\end{equation}
This co-training scheme couples the two roles through both rewards and data distributions.
The evaluation task provides denser and more outcome-aligned feedback for the search task by estimating how much each intermediate state supports the final answer.
At the same time, the search task supplies the evaluation task with states sampled from the current policy's own search distribution. This is important because an evaluator trained on states that differ from the policy's evolving search behavior may provide inaccurate process rewards. By optimizing search and evaluation rollouts with the same policy parameters, OASES keeps the evaluator adapted to the search policy while using the evaluator to improve search credit assignment. At inference time, only the search policy is executed, so OASES introduces no additional inference-time cost. Further discussion on the necessity of joint optimization is provided in Appendix~\ref{sec:joint_search_eval_discussion}.

\section{Experiments}

We conduct experiments to answer the following research questions:

\noindent \textbf{RQ1:} Does OASES consistently outperform existing baselines?

\noindent \textbf{RQ2:} Is joint search--evaluation optimization effective compared with non-co-trained policy or frozen evaluators?

\noindent \textbf{RQ3:} How much do outcome-aligned process rewards contribute?

\noindent \textbf{RQ4:} How sensitive is OASES to the reward weight $\alpha$?

\subsection{Experimental Settings}

\paragraph{Datasets.}

We train OASES on HotpotQA~\cite{yang2018hotpotqa} and 2WikiMultihopQA~\cite{ho2020constructing}, and evaluate it on NQ~\cite{kwiatkowski2019natural}, HotpotQA, 2WikiMultihopQA, Bamboogle~\cite{press2022measuring}, and MuSiQue~\cite{trivedi2022musique}.

\paragraph{Baselines.}

We compare OASES with three categories of baselines: (1) non-agentic methods, including direct LLM answering and naive RAG~\cite{lewis2020retrieval}; (2) agentic search methods, including Search-o1~\cite{li2025search}, Search-R1~\cite{jin2025search}, and R1-Searcher~\cite{song2025r1}; and (3) process-supervision methods, including StepSearch~\cite{zheng2025stepsearch}, ReasonRAG~\cite{zhang2025process}, TIPS~\cite{xie2026tips}, and SubSearch~\cite{petcu2026subsearch}.\footnote{For a fair comparison, we evaluate all baselines under the same training and inference settings as OASES. When available, we use released checkpoints for inference; otherwise, we reproduce the methods using their publicly released code within our training and evaluation pipeline, rather than directly quoting reported numbers.} Detailed descriptions of these baselines are provided in Appendix~\ref{app:baseline_details}.

\paragraph{Implementation Details.}

We implement OASES using \texttt{verl}\footnote{\url{https://github.com/volcengine/verl}} and train it with PPO. Unless otherwise specified, we initialize the policy from Qwen2.5-7B-Instruct~\cite{team2024qwen2}; we also evaluate OASES on other model sizes and model families, as reported in Appendix~\ref{app:qwen_base_results}. For retrieval, we use E5~\cite{wang2022text} as the retriever and Wikipedia as the retrieval corpus. We report exact match (EM) and token-level F1 as evaluation metrics, and set the process-reward weight $\alpha$ to $0.5$ by default. Environment details and training hyperparameters are provided in Appendix~\ref{app:implementation_details}, and the generated-token overhead of state-evaluation rollouts is analyzed in Appendix~\ref{app:efficiency}.

\subsection{Main Results Analysis (RQ1)}

\paragraph{OASES consistently improves agentic search performance.}

Table~\ref{tab:main_results} reports the main results on Qwen2.5-7B-Instruct. OASES achieves the best performance across all five benchmarks in both F1 and EM, outperforming standard RL methods, agentic search baselines, and process-supervision methods. On average, OASES improves over the strongest baseline by 4.08 F1 and 3.25 EM, with particularly clear gains on HotpotQA and 2WikiMultihopQA. Additional results across different model sizes and model families are provided in Appendix~\ref{app:qwen_base_results}.

\paragraph{Outcome-aligned state rewards provide more effective process supervision.}

Compared with outcome-only RL and agentic search baselines, OASES provides denser supervision by decomposing outcome improvement across intermediate search states, helping the policy identify useful reasoning and retrieval actions. It also outperforms prior process-supervision methods based on heuristics, external evaluators, or information gain, whose rewards may be only indirectly tied to the final outcome or rely on likelihood changes under partial evidence. In contrast, OASES evaluates each state via state-conditioned answer generation and scores it with the same verifiable outcome metric used for final answers, making its process rewards more outcome-aligned and adaptive to the evolving policy.

\begin{table*}[t]
\centering
\small
\setlength{\tabcolsep}{4.5pt}
\renewcommand{\arraystretch}{1.15}
\caption{Comprehensive QA benchmark results on Qwen2.5-7B-Instruct. \textbf{Bold} indicates the best performance. \textit{Impv. vs Best} denotes OASES's absolute improvement over the strongest baseline.}
\label{tab:main_results}
\resizebox{\linewidth}{!}{
\begin{tabular}{lcccccccccccc}
\toprule
\multirow{2}{*}{Method} 
& \multicolumn{2}{c}{NQ} 
& \multicolumn{2}{c}{HotpotQA} 
& \multicolumn{2}{c}{2Wiki} 
& \multicolumn{2}{c}{Bamboogle} 
& \multicolumn{2}{c}{MuSiQue} 
& \multicolumn{2}{c}{Avg.} \\
\cmidrule(lr){2-3} \cmidrule(lr){4-5} \cmidrule(lr){6-7} \cmidrule(lr){8-9} \cmidrule(lr){10-11} \cmidrule(lr){12-13}
& F1 & EM & F1 & EM & F1 & EM & F1 & EM & F1 & EM & F1 & EM \\
\midrule
\rowcolor{gray!12}
\multicolumn{13}{c}{\textit{Standard}} \\
Direct LLM 
& 17.16 & 5.98 
& 25.25 & 17.60 
& 25.80 & 22.03 
& 14.38 & 8.00 
& 10.14 & 2.57 
& 18.55 & 11.24 \\
Naive RAG~\cite{lewis2020retrieval}
& 37.26 & 18.59 
& 36.69 & 27.45 
& 27.06 & 21.98 
& 26.87 & 16.80 
& 9.32 & 3.52 
& 27.44 & 17.67 \\
\midrule
\rowcolor{gray!12}
\multicolumn{13}{c}{\textit{Agentic Search}} \\
Search-o1~\cite{li2025search}
& 32.43 & 15.71 
& 44.86 & 34.07 
& 39.69 & 32.71 
& 37.56 & 27.20 
& 20.11 & 11.09 
& 34.93 & 24.16 \\
Search-R1~\cite{jin2025search}
& 40.98 & 20.78 
& 51.24 & 39.77 
& 47.10 & 39.71 
& 48.43 & 36.54 
& 22.50 & 12.70 
& 42.05 & 29.90 \\
R1-Searcher~\cite{song2025r1}
& 41.15 & 21.61 
& 55.55 & 43.31 
& 50.64 & 43.97 
& 54.62 & 44.00 
& 27.74 & 18.16 
& 45.94 & 34.21 \\
\midrule
\rowcolor{gray!12}
\multicolumn{13}{c}{\textit{Process Reward}} \\
StepSearch~\cite{zheng2025stepsearch}
& 39.79 & 19.81 
& 50.02 & 38.31 
& 44.35 & 37.59 
& 50.32 & 39.20 
& 27.50 & 18.16 
& 42.40 & 30.61 \\
ReasonRAG~\cite{zhang2025process}
& 39.65 & 19.78 
& 41.53 & 31.24 
& 36.58 & 31.39 
& 40.53 & 28.80 
& 16.37 & 8.94 
& 34.93 & 24.03 \\
TIPS~\cite{xie2026tips}
& 40.38 & 20.30 
& 45.98 & 35.34 
& 42.95 & 36.51 
& 46.20 & 36.00 
& 18.59 & 10.59 
& 38.82 & 27.75 \\
SubSearch~\cite{petcu2026subsearch}
& 41.00 & 21.00
& 54.30 & 43.04
& 51.71 & 46.15
& 51.08 & 42.40
& 23.24 & 15.06
& 44.27 & 33.53 \\

\rowcolor{cyan!8}
\textbf{OASES (Ours)}
& \textbf{43.62} & \textbf{22.60} 
& \textbf{60.62} & \textbf{47.17} 
& \textbf{58.14} & \textbf{50.48} 
& \textbf{56.99} & \textbf{48.00} 
& \textbf{30.73} & \textbf{19.03} 
& \textbf{50.02} & \textbf{37.46} \\
\rowcolor{cyan!8}
\textit{Impv. vs Best}
& +2.47 & +0.99 
& +5.07 & +3.86 
& +6.43 & +4.33 
& +2.37 & +4.00 
& +2.99 & +0.87 
& +4.08 & +3.25 \\
\bottomrule
\end{tabular}
}
\end{table*}

\subsection{How Joint Optimization Shapes the Search-State Evaluator (RQ2)}

To answer \textbf{RQ2}, we examine whether search--evaluation co-training improves the model's ability to evaluate intermediate search states. We compare three settings: \textbf{OASES}, which jointly optimizes search and state evaluation with shared model parameters; \textbf{OASES w/o Joint Eval.}, which uses the current policy as the state evaluator but does not train on state-evaluation rollouts; and \textbf{OASES w/ Frozen Eval.}, which uses the initial policy as a frozen evaluator throughout training. To make the effect of co-training explicit, we also report \textit{Impv. vs Best}, defined as the F1 improvement of OASES over the stronger non-co-trained alternative at each training step, i.e., the higher F1 between \textbf{OASES w/o Joint Eval.} and \textbf{OASES w/ Frozen Eval.}.

\paragraph{The gains are largest at intermediate search states.}

\begin{figure*}[t]
    \centering
    \begin{minipage}{0.24\textwidth}
        \centering
        \includegraphics[width=\textwidth]{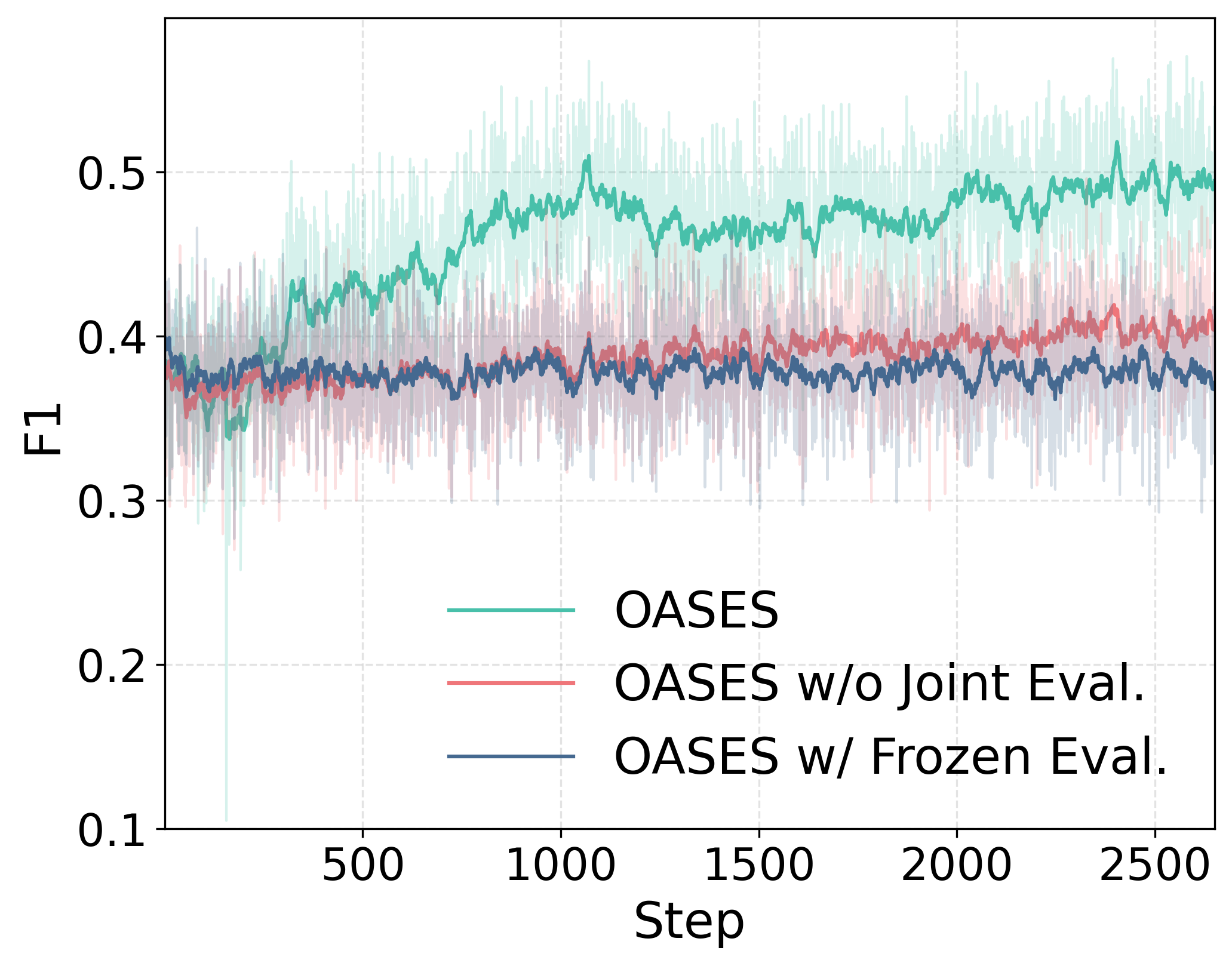}
        \small (a) Turn 0
    \end{minipage}
    \hfill
    \begin{minipage}{0.24\textwidth}
        \centering
        \includegraphics[width=\textwidth]{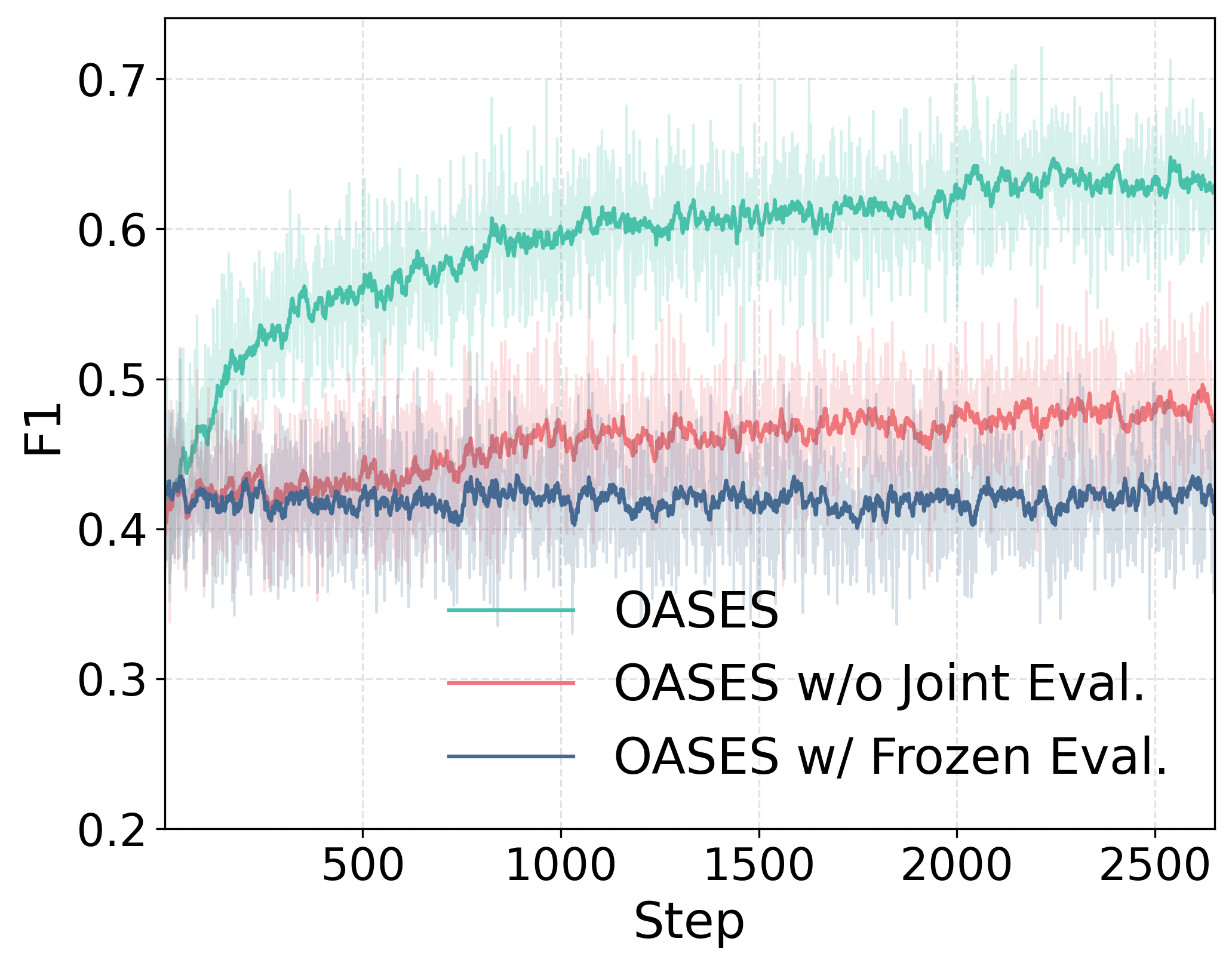}
        \small (b) Turn 1
    \end{minipage}
    \hfill
    \begin{minipage}{0.24\textwidth}
        \centering
        \includegraphics[width=\textwidth]{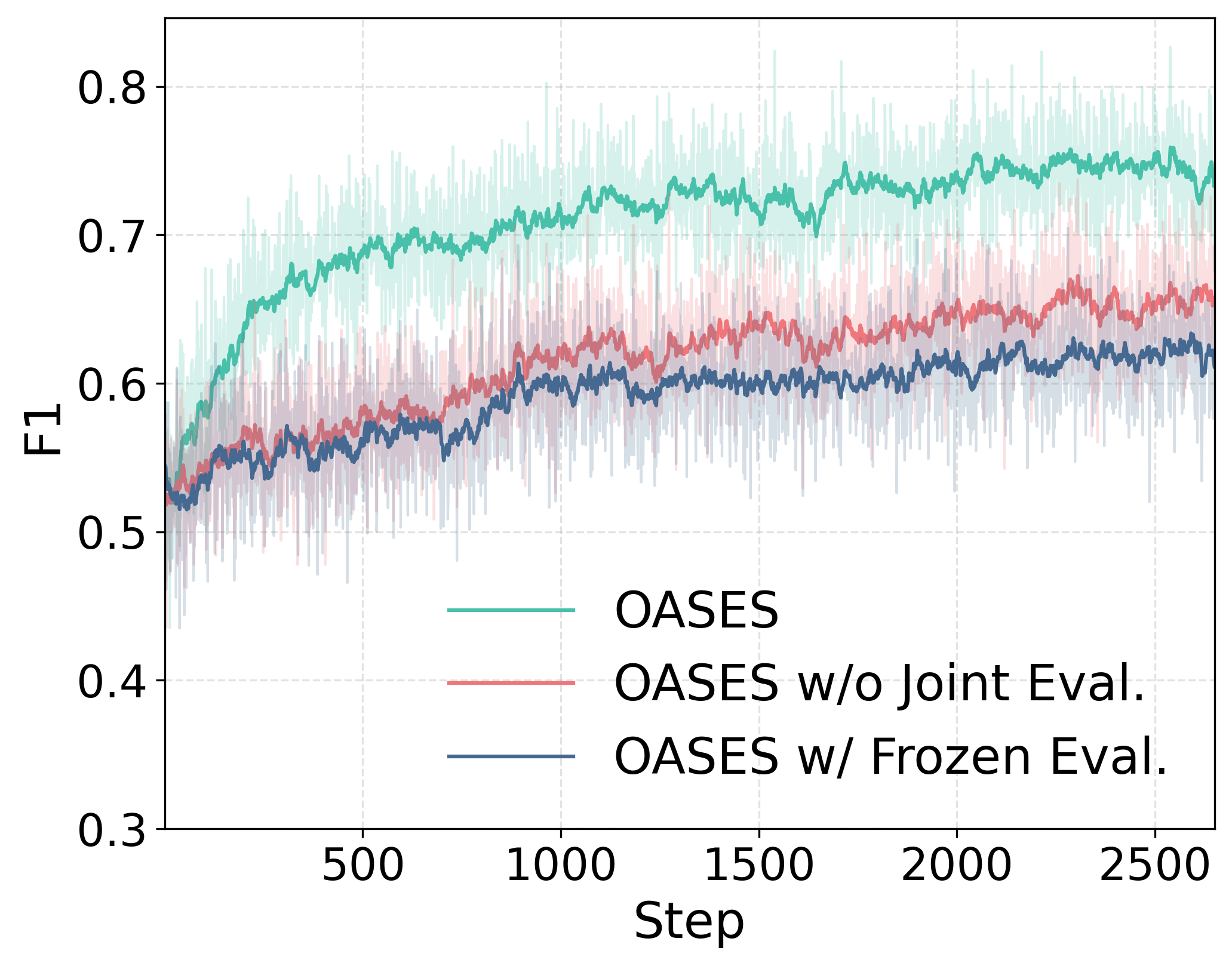}
        \small (c) Turn $\geq 2$
    \end{minipage}
    \hfill
    \begin{minipage}{0.24\textwidth}
        \centering
        \includegraphics[width=\textwidth]{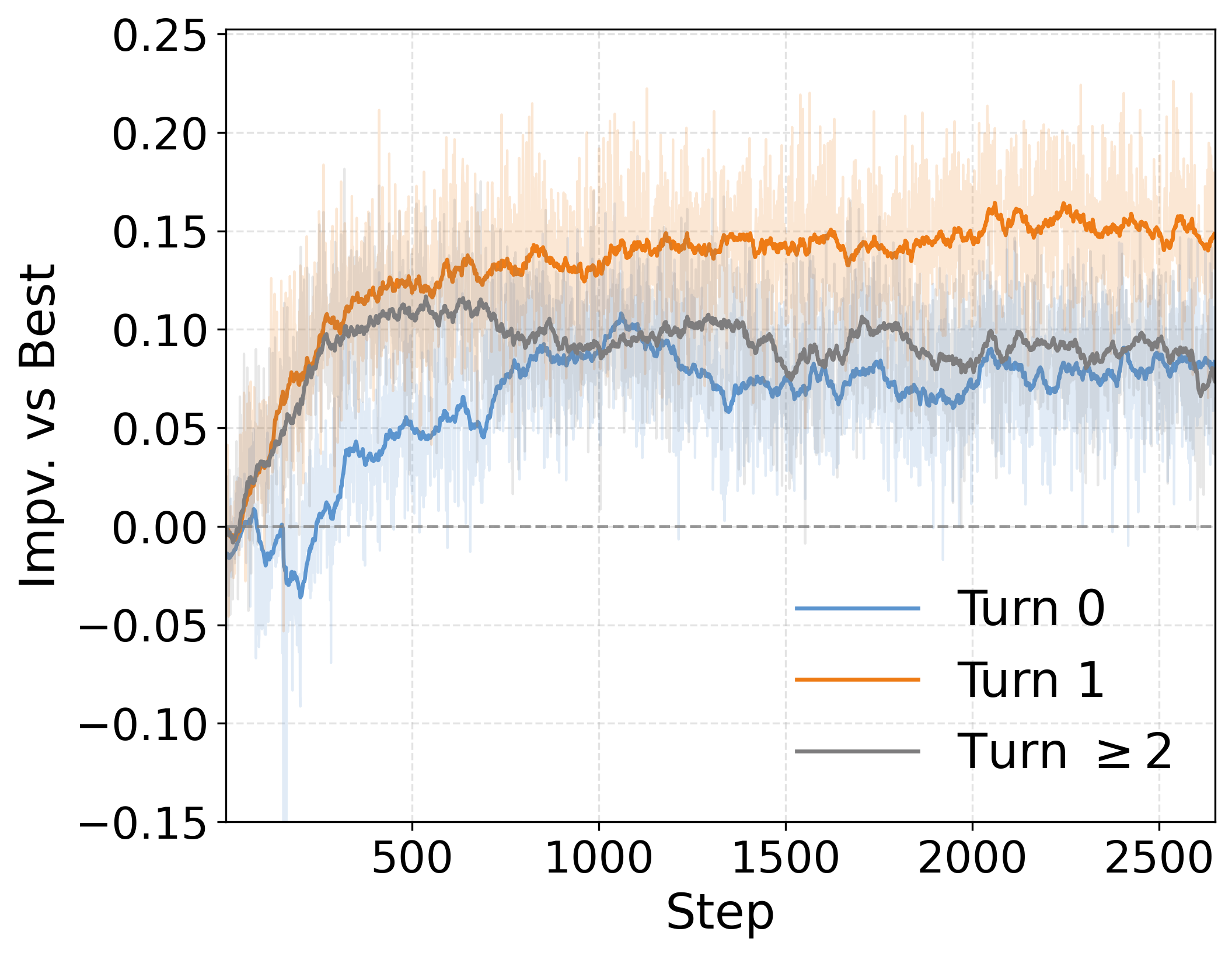}
        \small (d) Impv. vs Best
    \end{minipage}
    \caption{Step-wise analysis of the search-state evaluator under different optimization strategies. Panels (a)--(c) show the evaluator's answer F1 across training steps for search states at Turn~0, Turn~1, and Turn~$\geq 2$, respectively. Panel (d) reports \textit{Impv. vs Best}, defined as OASES's F1 improvement over the stronger non-joint setting at each turn.}
    \label{fig:joint_eval_analysis}
\end{figure*}

As shown in Figure~\ref{fig:joint_eval_analysis}, OASES consistently achieves the highest state-evaluation F1 across search stages. This indicates that search--evaluation co-training not only improves the search policy, but also strengthens the model's ability to judge whether its accumulated evidence supports the correct answer. The improvement is most pronounced at Turn~1. At Turn~0, the evaluator only observes the original question and has access to little external evidence, leaving limited room for improvement. At Turn~$\geq 2$, the accumulated evidence is often sufficient for answering, making evaluation easier for all variants. Turn~1 is the most informative regime: the state has acquired some useful evidence but remains incomplete, so accurate evaluation is crucial for assigning credit to the preceding search action. By jointly optimizing the evaluator with the evolving search policy, OASES provides more reliable feedback in this intermediate regime, while the non-joint and frozen evaluators suffer from weaker adaptation to the policy's changing search behavior.

\begin{figure*}[t]
\centering

\begin{minipage}[t]{0.50\textwidth}
\centering
\small
\setlength{\tabcolsep}{3.2pt}
\renewcommand{\arraystretch}{1.08}
\captionof{table}{Ablation results on three representative benchmarks. $\Delta$ denotes the performance difference compared with the full OASES model.}
\label{tab:ablation}
\resizebox{\linewidth}{!}{
\begin{tabular}{lcccccc}
\toprule
\multirow{2}{*}{Method}
& \multicolumn{2}{c}{HotpotQA}
& \multicolumn{2}{c}{2Wiki}
& \multicolumn{2}{c}{MuSiQue} \\
\cmidrule(lr){2-3} \cmidrule(lr){4-5} \cmidrule(lr){6-7}
& F1 & $\Delta$ & F1 & $\Delta$ & F1 & $\Delta$ \\
\midrule
\textbf{OASES}
& \textbf{60.62} & -
& \textbf{59.69} & -
& \textbf{30.73} & - \\
\midrule
\multicolumn{7}{l}{\textit{w/o joint opt.}} \\
Policy model
& 57.16 & -3.46
& 57.81 & -1.88
& 27.46 & -3.27 \\
Frozen Qwen-7B
& 57.65 & -2.97
& 57.93 & -1.76
& 27.88 & -2.85 \\
Frozen Qwen-14B
& 57.57 & -3.05
& 58.85 & -0.84
& 26.93 & -3.80 \\
Frozen Qwen-32B
& 57.11 & -3.51
& 58.51 & -1.18
& 28.05 & -2.68 \\
Frozen GPT-4o
& 57.38 & -3.24
& 58.82 & -0.87
& 28.24 & -2.49 \\
\midrule
\textit{w/o proc. reward}
& 58.80 & -1.82
& 58.64 & -1.05
& 27.62 & -3.11 \\
\textit{w/o joint opt. \& proc. reward}
& 58.42 & -2.20
& 58.00 & -1.69
& 28.79 & -1.94 \\
\bottomrule
\end{tabular}
}
\end{minipage}
\hfill
\begin{minipage}[t]{0.47\textwidth}
\centering
\vspace{0pt}

\begin{minipage}{0.49\linewidth}
    \centering
    \includegraphics[width=\linewidth]{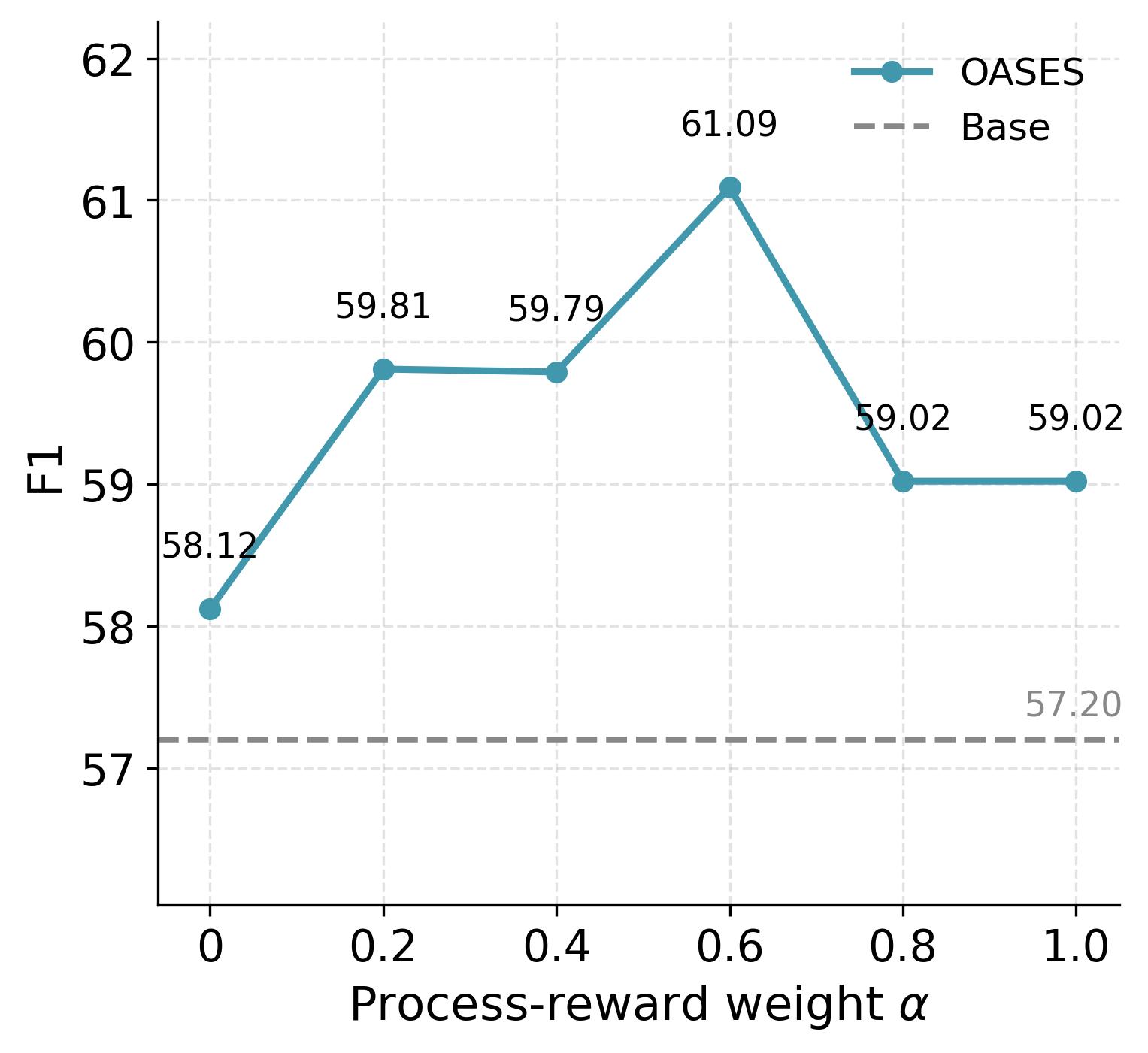}\\[-1mm]
    {\small (a) 7B model on F1}
\end{minipage}
\hfill
\begin{minipage}{0.49\linewidth}
    \centering
    \includegraphics[width=\linewidth]{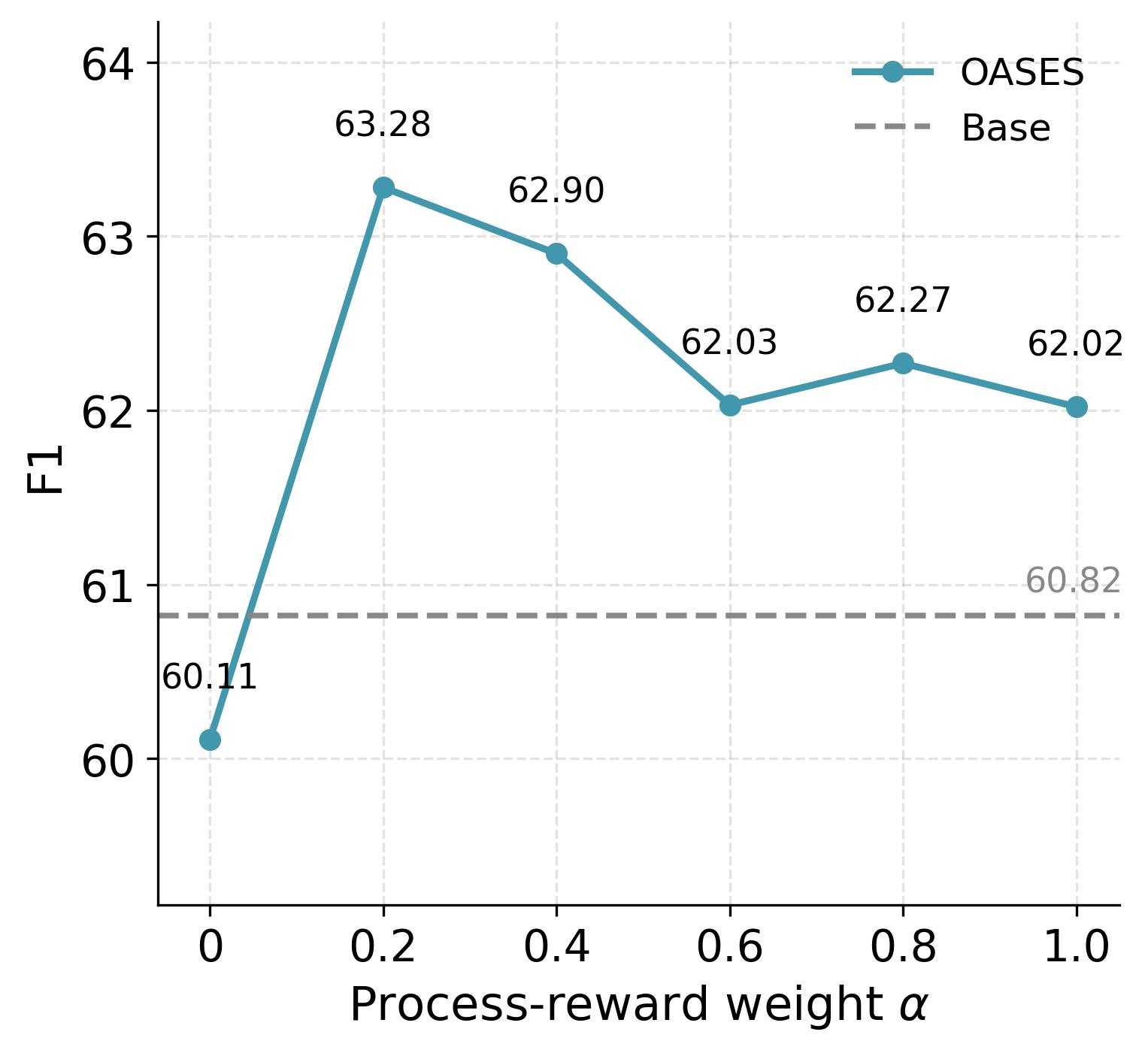}\\[-1mm]
    {\small (b) 14B model on F1}
\end{minipage}

\captionof{figure}{Effect of the process-reward weight $\alpha$ across model sizes. Panels (a) and (b) show F1 results for 7B and 14B. \textbf{Base} removes both process rewards and joint optimization.}
\label{fig:alpha_ablation}
\end{minipage}

\end{figure*}

\subsection{Ablation Study on Search--Evaluation Co-Training and Process Rewards (RQ2 \& RQ3)}

Table~\ref{tab:ablation} reports ablation results on three representative benchmarks. \textbf{OASES} is the full framework. Under \textbf{OASES w/o joint opt.}, we disable search--evaluation co-training and compare different choices of state evaluator, including the policy model itself and several frozen evaluators. \textbf{OASES w/o proc. reward} removes the state-level process rewards by setting $\alpha=0$. \textbf{OASES w/o joint opt. \& proc. reward} removes both co-training and process rewards, corresponding to the base variant.

\paragraph{Frozen or non-co-trained evaluators are insufficient.}

OASES performs best across all three benchmarks, showing the importance of both search--evaluation co-training and outcome-aligned process rewards. Removing joint optimization consistently degrades performance, even with stronger frozen evaluators such as GPT-4o, suggesting that evaluator scale alone cannot replace adaptation to the evolving search policy. Using the policy model as an evaluator without co-training is also suboptimal, since RL may specialize the model for search while weakening standalone state evaluation. In contrast, OASES jointly optimizes search and evaluation with shared parameters, keeping process supervision aligned with the evolving policy. The drop after removing process rewards further shows that state-level feedback provides useful supervision beyond terminal rewards. A qualitative failure case of search-only training is provided in Appendix~\ref{app:search_only_case}. 

\subsection{Effect of the Process-Reward Weight (RQ4)}

Figure~\ref{fig:alpha_ablation} shows the effect of the process-reward weight $\alpha$ on F1 across model sizes. The experiment is conducted on 2WikiMultihopQA with both 7B and 14B models. \textbf{Base} denotes the variant without both outcome-aligned process rewards and search--evaluation co-training. The setting $\alpha=0$ keeps co-training but removes the contribution of process rewards, while positive values of $\alpha$ assign different weights to state-level outcome improvements.

\paragraph{Outcome-aligned feedback is consistently beneficial.}

The results show that \textbf{Base} and $\alpha=0$ are generally worse than using a positive process-reward weight, confirming the importance of dense outcome-aligned feedback. For the 7B model, performance peaks at $\alpha=0.6$, while for the 14B model, the best result is achieved at $\alpha=0.2$. Although the optimal value differs across model sizes, positive values of $\alpha$ consistently outperform the no-process-reward setting, showing that OASES is robust to the choice of $\alpha$ within a reasonable range.

\section{Conclusion}

We presented OASES, an outcome-aligned search--evaluation co-training framework for agentic search. OASES uses the same verifiable outcome signal for state-level and final-answer evaluation, converting outcome improvements across search states into dense process rewards. By co-training a single policy for search and state evaluation, OASES provides outcome-aligned credit assignment while reducing evaluator--policy mismatch. Experiments on multi-hop QA benchmarks show that OASES consistently outperforms strong RL baselines, and ablations confirm the importance of outcome-aligned process rewards and search--evaluation co-training.

\bibliographystyle{plainnat}
\bibliography{references}

@article{jin2025search,
  title={Search-r1: Training llms to reason and leverage search engines with reinforcement learning},
  author={Jin, Bowen and Zeng, Hansi and Yue, Zhenrui and Yoon, Jinsung and Arik, Sercan and Wang, Dong and Zamani, Hamed and Han, Jiawei},
  journal={arXiv preprint arXiv:2503.09516},
  year={2025}
}

@article{song2025r1,
  title={R1-searcher: Incentivizing the search capability in llms via reinforcement learning},
  author={Song, Huatong and Jiang, Jinhao and Min, Yingqian and Chen, Jie and Chen, Zhipeng and Zhao, Wayne Xin and Fang, Lei and Wen, Ji-Rong},
  journal={arXiv preprint arXiv:2503.05592},
  year={2025}
}

@article{lewis2020retrieval,
  title={Retrieval-augmented generation for knowledge-intensive nlp tasks},
  author={Lewis, Patrick and Perez, Ethan and Piktus, Aleksandra and Petroni, Fabio and Karpukhin, Vladimir and Goyal, Naman and K{\"u}ttler, Heinrich and Lewis, Mike and Yih, Wen-tau and Rockt{\"a}schel, Tim and others},
  journal={Advances in neural information processing systems},
  volume={33},
  pages={9459--9474},
  year={2020}
}

@inproceedings{li2025search,
  title={Search-o1: Agentic search-enhanced large reasoning models},
  author={Li, Xiaoxi and Dong, Guanting and Jin, Jiajie and Zhang, Yuyao and Zhou, Yujia and Zhu, Yutao and Zhang, Peitian and Dou, Zhicheng},
  booktitle={Proceedings of the 2025 Conference on Empirical Methods in Natural Language Processing},
  pages={5420--5438},
  year={2025}
}

@inproceedings{yao2022react,
  title={React: Synergizing reasoning and acting in language models},
  author={Yao, Shunyu and Zhao, Jeffrey and Yu, Dian and Du, Nan and Shafran, Izhak and Narasimhan, Karthik R and Cao, Yuan},
  booktitle={The eleventh international conference on learning representations},
  year={2022}
}

@article{schick2023toolformer,
  title={Toolformer: Language models can teach themselves to use tools},
  author={Schick, Timo and Dwivedi-Yu, Jane and Dess{\`\i}, Roberto and Raileanu, Roberta and Lomeli, Maria and Hambro, Eric and Zettlemoyer, Luke and Cancedda, Nicola and Scialom, Thomas},
  journal={Advances in neural information processing systems},
  volume={36},
  pages={68539--68551},
  year={2023}
}

@article{lambert2024tulu,
  title={Tulu 3: Pushing frontiers in open language model post-training},
  author={Lambert, Nathan and Morrison, Jacob and Pyatkin, Valentina and Huang, Shengyi and Ivison, Hamish and Brahman, Faeze and Miranda, Lester James V and Liu, Alisa and Dziri, Nouha and Lyu, Shane and others},
  journal={arXiv preprint arXiv:2411.15124},
  year={2024}
}

@article{guo2025deepseek,
  title={Deepseek-r1: Incentivizing reasoning capability in llms via reinforcement learning},
  author={Guo, Daya and Yang, Dejian and Zhang, Haowei and Song, Junxiao and Wang, Peiyi and Zhu, Qihao and Xu, Runxin and Zhang, Ruoyu and Ma, Shirong and Bi, Xiao and others},
  journal={arXiv preprint arXiv:2501.12948},
  year={2025}
}

@inproceedings{zheng2025stepsearch,
  title={StepSearch: Igniting LLMs search ability via step-wise proximal policy optimization},
  author={Zheng, Xuhui and An, Kang and Wang, Ziliang and Wang, Yuhang and Wu, Yichao},
  booktitle={Proceedings of the 2025 Conference on Empirical Methods in Natural Language Processing},
  pages={21816--21841},
  year={2025}
}

@article{cui2025process,
  title={Process reinforcement through implicit rewards},
  author={Cui, Ganqu and Yuan, Lifan and Wang, Zefan and Wang, Hanbin and Zhang, Yuchen and Chen, Jiacheng and Li, Wendi and He, Bingxiang and Fan, Yuchen and Yu, Tianyu and others},
  journal={arXiv preprint arXiv:2502.01456},
  year={2025}
}

@inproceedings{xie2026tips,
  title={TIPS: Turn-level Information-Potential Reward Shaping for Search-Augmented LLMs},
  author={Xie, Yutao and Thomas, Nathaniel and Hansen, Nicklas and Fu, Yang and Li, Li Erran and Wang, Xiaolong},
  booktitle={The Fourteenth International Conference on Learning Representations},
  year={2026}
}

@article{zhang2025process,
  title={Process vs. outcome reward: Which is better for agentic RAG reinforcement learning},
  author={Zhang, Wenlin and Li, Xiangyang and Dong, Kuicai and Wang, Yichao and Jia, Pengyue and Li, Xiaopeng and Zhang, Yingyi and Xu, Derong and Du, Zhaocheng and Guo, Huifeng and others},
  journal={arXiv preprint arXiv:2505.14069},
  year={2025}
}

@article{chen2026jade,
  title={JADE: Bridging the Strategic-Operational Gap in Dynamic Agentic RAG},
  author={Chen, Yiqun and Zhang, Erhan and Hu, Tianyi and Wang, Shijie and Yang, Zixuan and Zhong, Meizhi and Wei, Xiaochi and Gao, Yan and Wu, Yi and Hu, Yao and others},
  journal={arXiv preprint arXiv:2601.21916},
  year={2026}
}

@article{chen2026beyond,
  title={Beyond Monolithic Architectures: A Multi-Agent Search and Knowledge Optimization Framework for Agentic Search},
  author={Chen, Yiqun and Yan, Lingyong and Yang, Zixuan and Zhang, Erhan and Zhao, Jiashu and Wang, Shuaiqiang and Yin, Dawei and Mao, Jiaxin},
  journal={arXiv preprint arXiv:2601.04703},
  year={2026}
}

@article{schulman2017proximal,
  title={Proximal policy optimization algorithms},
  author={Schulman, John and Wolski, Filip and Dhariwal, Prafulla and Radford, Alec and Klimov, Oleg},
  journal={arXiv preprint arXiv:1707.06347},
  year={2017}
}

@article{shao2024deepseekmath,
  title={Deepseekmath: Pushing the limits of mathematical reasoning in open language models},
  author={Shao, Zhihong and Wang, Peiyi and Zhu, Qihao and Xu, Runxin and Song, Junxiao and Bi, Xiao and Zhang, Haowei and Zhang, Mingchuan and Li, YK and Wu, Yang and others},
  journal={arXiv preprint arXiv:2402.03300},
  year={2024}
}

@article{chen2025mao,
  title={Mao-arag: Multi-agent orchestration for adaptive retrieval-augmented generation},
  author={Chen, Yiqun and Zhang, Erhan and Yan, Lingyong and Wang, Shuaiqiang and Huang, Jizhou and Yin, Dawei and Mao, Jiaxin},
  journal={arXiv preprint arXiv:2508.01005},
  year={2025}
}

@article{ma2023query,
  title={Query rewriting for retrieval-augmented large language models},
  author={Ma, Xinbei and Gong, Yeyun and He, Pengcheng and Zhao, Hai and Duan, Nan},
  journal={arXiv preprint arXiv:2305.14283},
  year={2023}
}

@article{li2024dmqr,
  title={Dmqr-rag: Diverse multi-query rewriting for rag},
  author={Li, Zhicong and Wang, Jiahao and Jiang, Zhishu and Mao, Hangyu and Chen, Zhongxia and Du, Jiazhen and Zhang, Yuanxing and Zhang, Fuzheng and Zhang, Di and Liu, Yong},
  journal={arXiv preprint arXiv:2411.13154},
  year={2024}
}

@article{ke2024bridging,
  title={Bridging the preference gap between retrievers and llms},
  author={Ke, Zixuan and Kong, Weize and Li, Cheng and Zhang, Mingyang and Mei, Qiaozhu and Bendersky, Michael},
  journal={arXiv preprint arXiv:2401.06954},
  year={2024}
}

@article{li2024rag,
  title={RAG-DDR: Optimizing Retrieval-Augmented Generation Using Differentiable Data Rewards},
  author={Li, Xinze and Mei, Sen and Liu, Zhenghao and Yan, Yukun and Wang, Shuo and Yu, Shi and Zeng, Zheni and Chen, Hao and Yu, Ge and Liu, Zhiyuan and others},
  journal={arXiv preprint arXiv:2410.13509},
  year={2024}
}

@article{yang2025qwen3,
  title={Qwen3 technical report},
  author={Yang, An and Li, Anfeng and Yang, Baosong and Zhang, Beichen and Hui, Binyuan and Zheng, Bo and Yu, Bowen and Gao, Chang and Huang, Chengen and Lv, Chenxu and others},
  journal={arXiv preprint arXiv:2505.09388},
  year={2025}
}

@article{wang2025information,
  title={Information Gain-based Policy Optimization: A Simple and Effective Approach for Multi-Turn LLM Agents},
  author={Wang, Guoqing and Dai, Sunhao and Ye, Guangze and Gan, Zeyu and Yao, Wei and Deng, Yong and Wu, Xiaofeng and Ying, Zhenzhe},
  journal={arXiv preprint arXiv:2510.14967},
  year={2025}
}

@inproceedings{lightman2023let,
  title={Let's verify step by step},
  author={Lightman, Hunter and Kosaraju, Vineet and Burda, Yuri and Edwards, Harrison and Baker, Bowen and Lee, Teddy and Leike, Jan and Schulman, John and Sutskever, Ilya and Cobbe, Karl},
  booktitle={The twelfth international conference on learning representations},
  year={2023}
}

@article{chen2025improving,
  title={Improving retrieval-augmented generation through multi-agent reinforcement learning},
  author={Chen, Yiqun and Yan, Lingyong and Sun, Weiwei and Ma, Xinyu and Zhang, Yi and Wang, Shuaiqiang and Yin, Dawei and Yang, Yiming and Mao, Jiaxin},
  journal={arXiv preprint arXiv:2501.15228},
  year={2025}
}

@article{kwiatkowski2019natural,
  title={Natural questions: a benchmark for question answering research},
  author={Kwiatkowski, Tom and Palomaki, Jennimaria and Redfield, Olivia and Collins, Michael and Parikh, Ankur and Alberti, Chris and Epstein, Danielle and Polosukhin, Illia and Devlin, Jacob and Lee, Kenton and others},
  journal={Transactions of the Association for Computational Linguistics},
  volume={7},
  pages={453--466},
  year={2019},
  publisher={MIT Press One Rogers Street, Cambridge, MA 02142-1209, USA journals-info~…}
}

@article{yang2018hotpotqa,
  title={HotpotQA: A dataset for diverse, explainable multi-hop question answering},
  author={Yang, Zhilin and Qi, Peng and Zhang, Saizheng and Bengio, Yoshua and Cohen, William W and Salakhutdinov, Ruslan and Manning, Christopher D},
  journal={arXiv preprint arXiv:1809.09600},
  year={2018}
}

@article{ho2020constructing,
  title={Constructing a multi-hop QA dataset for comprehensive evaluation of reasoning steps},
  author={Ho, Xanh and Nguyen, Anh-Khoa Duong and Sugawara, Saku and Aizawa, Akiko},
  journal={arXiv preprint arXiv:2011.01060},
  year={2020}
}

@article{press2022measuring,
  title={Measuring and narrowing the compositionality gap in language models},
  author={Press, Ofir and Zhang, Muru and Min, Sewon and Schmidt, Ludwig and Smith, Noah A and Lewis, Mike},
  journal={arXiv preprint arXiv:2210.03350},
  year={2022}
}

@article{trivedi2022musique,
  title={MuSiQue: Multihop Questions via Single-hop Question Composition},
  author={Trivedi, Harsh and Balasubramanian, Niranjan and Khot, Tushar and Sabharwal, Ashish},
  journal={Transactions of the Association for Computational Linguistics},
  volume={10},
  pages={539--554},
  year={2022},
  publisher={MIT Press One Broadway, 12th Floor, Cambridge, Massachusetts 02142, USA~…}
}

@article{wang2022text,
  title={Text embeddings by weakly-supervised contrastive pre-training},
  author={Wang, Liang and Yang, Nan and Huang, Xiaolong and Jiao, Binxing and Yang, Linjun and Jiang, Daxin and Majumder, Rangan and Wei, Furu},
  journal={arXiv preprint arXiv:2212.03533},
  year={2022}
}

@article{team2024qwen2,
  title={Qwen2.5 Technical Report},
  author={Team, Qwen},
  journal={arXiv preprint arXiv:2412.15115},
  year={2024}
}

@article{wang2026not,
  title={Do Not Waste Your Rollouts: Recycling Search Experience for Efficient Test-Time Scaling},
  author={Wang, Xinglin and Shi, Jiayi and Feng, Shaoxiong and Yuan, Peiwen and Li, Yiwei and Zhang, Yueqi and Tan, Chuyi and Zhang, Ji and Pan, Boyuan and Hu, Yao and others},
  journal={arXiv preprint arXiv:2601.21684},
  year={2026}
}

@article{petcu2026subsearch,
  title={SubSearch: Intermediate Rewards for Unsupervised Guided Reasoning in Complex Retrieval},
  author={Petcu, Roxana and Kanoulas, Evangelos and de Rijke, Maarten},
  journal={arXiv preprint arXiv:2604.07415},
  year={2026}
}

@article{liang2026ig,
  title={IG-Search: Step-Level Information Gain Rewards for Search-Augmented Reasoning},
  author={Liang, Zihan and Ma, Yufei and Chen, Ben and Qian, Zhipeng and Dai, Huangyu and Mao, Lingtao and Zhang, Xuxin and Lei, Chenyi and Ou, Wenwu},
  journal={arXiv preprint arXiv:2604.15148},
  year={2026}
}

@article{xia2026search,
  title={Search-P1: Path-Centric Reward Shaping for Stable and Efficient Agentic RAG Training},
  author={Xia, Tianle and Xu, Ming and Hu, Lingxiang and Sun, Yiding and Li, Wenwei and Shang, Linfang and Liu, Liqun and Shu, Peng and Yu, Huan and Jiang, Jie},
  journal={arXiv preprint arXiv:2602.22576},
  year={2026}
}

@article{tan2025process,
  title={Process-Supervised Reinforcement Learning for Interactive Multimodal Tool-Use Agents},
  author={Tan, Weiting and Qu, Xinghua and Tu, Ming and Ge, Meng and Liu, Andy T and Koehn, Philipp and Lu, Lu},
  journal={arXiv preprint arXiv:2509.14480},
  year={2025}
}

@article{wang2026prorag,
  title={ProRAG: Process-Supervised Reinforcement Learning for Retrieval-Augmented Generation},
  author={Wang, Zhao and Zhao, Ziliang and Dou, Zhicheng},
  journal={arXiv preprint arXiv:2601.21912},
  year={2026}
}

@article{xuprinciple,
  title={Principle Process Reward For Search Agents},
  author={Xu, Peiran and Li, Zhuohao and Xing, Xiaoying and Zhang, Guannan and Li, Debiao and Shi, Kunyu}
}

@article{zhao2025r,
  title={R-search: Empowering llm reasoning with search via multi-reward reinforcement learning},
  author={Zhao, Qingfei and Wang, Ruobing and Xu, Dingling and Zha, Daren and Liu, Limin},
  journal={arXiv preprint arXiv:2506.04185},
  year={2025}
}

@article{wang2026enhancing,
  title={Enhancing LLM-based Search Agents via Contribution Weighted Group Relative Policy Optimization},
  author={Wang, Junzhe and Xi, Zhiheng and Luo, Hao and Dou, Shihan and Gui, Tao and Zhang, Qi and others},
  journal={arXiv preprint arXiv:2604.14267},
  year={2026}
}

@article{shi2025search,
  title={Search and Refine During Think: Facilitating Knowledge Refinement for Improved Retrieval-Augmented Reasoning},
  author={Shi, Yaorui and Li, Sihang and Wu, Chang and Liu, Zhiyuan and Fang, Junfeng and Cai, Hengxing and Zhang, An and Wang, Xiang},
  journal={arXiv preprint arXiv:2505.11277},
  year={2025}
}

@article{lin2025comprehensive,
  title={A comprehensive survey on reinforcement learning-based agentic search: Foundations, roles, optimizations, evaluations, and applications},
  author={Lin, Minhua and Wu, Zongyu and Xu, Zhichao and Liu, Hui and Tang, Xianfeng and He, Qi and Aggarwal, Charu and Zhang, Xiang and Wang, Suhang},
  journal={arXiv preprint arXiv:2510.16724},
  year={2025}
}

@article{hu2026optimizing,
  title={Optimizing Agentic Reasoning with Retrieval via Synthetic Semantic Information Gain Reward},
  author={Hu, Senkang and Dai, Yong and Zhao, Yuzhi and Tao, Yihang and Guo, Yu and Fang, Zhengru and Kwong, Sam Tak Wu and Fang, Yuguang},
  journal={arXiv preprint arXiv:2602.00845},
  year={2026}
}

\newpage

\appendix

\section{Additional Results across Model Sizes and Model Families}
\label{app:qwen_base_results}

We provide additional experiments to complement the main Qwen2.5-7B-Instruct results. Table~\ref{tab:model_size_results} evaluates OASES on smaller Qwen2.5-Instruct models. OASES consistently improves over agentic search baselines on both Qwen2.5-1.5B-Instruct and Qwen2.5-3B-Instruct, showing that outcome-aligned search--evaluation co-training is also effective under smaller model scales. The gains are especially pronounced for Qwen2.5-1.5B-Instruct, where OASES improves the average F1 from 22.84 with Search-R1 to 42.55.

Table~\ref{tab:qwen_base_results} further reports results on Qwen2.5 and Qwen3 base models before OASES training. These results provide an additional reference for understanding how model family and scale affect agentic search performance under the base-model setting. Qwen3-8B-Base achieves the strongest performance among the evaluated base models, suggesting that stronger base capabilities can provide a better starting point for agentic search training.

\begin{table*}[h]
\centering
\small
\setlength{\tabcolsep}{4.5pt}
\renewcommand{\arraystretch}{1.15}
\caption{Additional QA benchmark results on smaller Qwen2.5-Instruct models. \textbf{Bold} indicates best performance within each model-size group.}
\label{tab:model_size_results}
\resizebox{\linewidth}{!}{
\begin{tabular}{lcccccccccccc}
\toprule
\multirow{2}{*}{Method} 
& \multicolumn{2}{c}{NQ} 
& \multicolumn{2}{c}{HotpotQA} 
& \multicolumn{2}{c}{2Wiki} 
& \multicolumn{2}{c}{Bamboogle} 
& \multicolumn{2}{c}{MuSiQue} 
& \multicolumn{2}{c}{Avg.} \\
\cmidrule(lr){2-3} \cmidrule(lr){4-5} \cmidrule(lr){6-7} \cmidrule(lr){8-9} \cmidrule(lr){10-11} \cmidrule(lr){12-13}
& F1 & EM & F1 & EM & F1 & EM & F1 & EM & F1 & EM & F1 & EM \\
\midrule
\rowcolor{gray!12}
\multicolumn{13}{c}{\textit{Qwen2.5-1.5B-Instruct}} \\

Search-o1~\cite{li2025search}
& 1.23 & 0.69
& 4.51 & 3.43
& 4.82 & 4.13
& 4.83 & 3.20
& 1.36 & 0.54
& 3.35 & 2.40 \\
Search-R1~\cite{jin2025search}
& 31.72 & 15.96
& 31.24 & 22.67
& 25.73 & 21.60
& 15.11 & 6.73
& 10.40 & 3.89
& 22.84 & 14.17 \\
SubSearch~\cite{petcu2026subsearch}
& 17.36 & 6.81
& 22.08 & 15.33
& 25.72 & 21.88
& 28.04 & 18.40
& 8.16 & 2.36
& 20.27 & 12.96 \\
\textbf{OASES (Ours)}
& \textbf{37.73} & \textbf{18.75}
& \textbf{51.39} & \textbf{38.69}
& \textbf{54.41} & \textbf{46.56}
& \textbf{46.41} & \textbf{37.60}
& \textbf{22.81} & \textbf{12.58}
& \textbf{42.55} & \textbf{30.84} \\
\midrule
\rowcolor{gray!12}
\multicolumn{13}{c}{\textit{Qwen2.5-3B-Instruct}} \\

Search-o1~\cite{li2025search}
& 26.36 & 12.22
& 33.27 & 24.34
& 28.46 & 23.40
& 38.66 & 28.80
& 13.34 & 5.50
& 28.02 & 18.85 \\
Search-R1~\cite{jin2025search}
& 35.70 & 17.95
& 40.61 & 31.57
& 41.83 & 36.74
& 31.12 & 21.15
& 17.07 & 9.39
& 33.27 & 23.36 \\
SubSearch~\cite{petcu2026subsearch}
& 39.05 & 19.78
& 50.73 & 39.41
& 51.00 & 45.09
& \textbf{51.55} & 39.20
& 24.12 & \textbf{15.06}
& 43.29 & 31.71 \\
\textbf{OASES (Ours)}
& \textbf{40.78} & \textbf{20.19}
& \textbf{55.31} & \textbf{42.81}
& \textbf{57.07} & \textbf{50.46}
& 51.53 & \textbf{40.80}
& \textbf{25.18} & 14.65
& \textbf{45.97} & \textbf{33.78} \\
\bottomrule
\end{tabular}
}
\end{table*}

\begin{table*}[h]
\centering
\small
\setlength{\tabcolsep}{4.5pt}
\renewcommand{\arraystretch}{1.15}
\caption{QA benchmark results on Qwen2.5 and Qwen3 base models with different model sizes.}
\label{tab:qwen_base_results}
\resizebox{\linewidth}{!}{
\begin{tabular}{lcccccccccccc}
\toprule
\multirow{2}{*}{Method} 
& \multicolumn{2}{c}{NQ} 
& \multicolumn{2}{c}{HotpotQA} 
& \multicolumn{2}{c}{2Wiki} 
& \multicolumn{2}{c}{Bamboogle} 
& \multicolumn{2}{c}{MuSiQue} 
& \multicolumn{2}{c}{Avg.} \\
\cmidrule(lr){2-3} \cmidrule(lr){4-5} \cmidrule(lr){6-7} \cmidrule(lr){8-9} \cmidrule(lr){10-11} \cmidrule(lr){12-13}
& F1 & EM & F1 & EM & F1 & EM & F1 & EM & F1 & EM & F1 & EM \\
\midrule
Qwen2.5-3B-Base
& 20.72 & 7.45
& 30.24 & 19.82
& 33.82 & 28.55
& 29.38 & 17.60
& 13.14 & 2.90
& 25.46 & 15.26 \\

Qwen3-4B-Base
& 20.89 & 6.93
& 30.80 & 19.86
& 34.36 & 29.17
& 17.55 & 8.00
& 13.79 & 2.57
& 23.48 & 11.73 \\

Qwen2.5-7B-Base
& 43.31 & 22.44
& 59.79 & 46.29
& 60.34 & 54.60
& 61.06 & 46.40
& 28.99 & 17.17
& 50.70 & 37.38 \\

Qwen3-8B-Base
& 44.08 & 22.94
& 60.79 & 47.52
& 61.73 & 55.66
& 61.15 & 49.60
& 31.92 & 20.56
& 51.93 & 39.26 \\
\bottomrule
\end{tabular}
}
\end{table*}

\section{Analysis of Outcome-Aligned Process Rewards}
\label{app:theory}

We provide an interpretation of OASES from the perspective of reward redistribution. Existing process-reward methods often introduce auxiliary reward sources whose optimization targets may deviate from the final task objective. In contrast, OASES derives dense process rewards by measuring how much each search step improves the model's ability to answer the original question under the same verifiable objective used for final-answer evaluation. Therefore, the process rewards can be viewed as an outcome-aligned decomposition of answerability improvements across intermediate search states. Since the state-level gain is defined as $g_t=\gamma_t-\gamma_{t-1}$, the accumulated process reward over a trajectory satisfies
\begin{equation}
\sum_{t=1}^{T} r_t^{\mathrm{proc}}
=
\alpha \sum_{t=1}^{T} (\gamma_t-\gamma_{t-1})
=
\alpha (\gamma_T-\gamma_0).
\label{eq:telescoping}
\end{equation}
Thus, OASES's process reward is not an independent reward source, but a telescoping redistribution of the improvement from the initial search state to the final search state. When the final search state provides sufficient evidence for answering the question, $\gamma_T$ is expected to correlate with the terminal outcome score $\gamma_{\mathrm{out}}$, and the trajectory-level reward can be interpreted as
\begin{equation}
G^{\mathrm{OASES}}(\tau)
=
\gamma_{\mathrm{out}}+\sum_{t=1}^{T} r_t^{\mathrm{proc}}
=
\gamma_{\mathrm{out}}+\alpha(\gamma_T-\gamma_0).
\label{eq:OASES_total_reward}
\end{equation}
Under the idealized case where the state-evaluation score at the final search state matches the final-answer score, i.e., $\gamma_T \approx \gamma_{\mathrm{out}}$, this becomes
\begin{equation}
G^{\mathrm{OASES}}(\tau)
\approx
\gamma_{\mathrm{out}}+\alpha(\gamma_{\mathrm{out}}-\gamma_0).
\end{equation}
In the special case where $\alpha=1$, this further reduces to
\begin{equation}
G^{\mathrm{OASES}}(\tau)
\approx
2\gamma_{\mathrm{out}}-\gamma_0.
\end{equation}
This analysis suggests that OASES encourages trajectories whose search process improves the answerability of intermediate states under the same verifiable scoring function used for final answers. Rather than introducing a separate proxy target, the dense process reward redistributes state-level answerability improvements back to intermediate search decisions, providing step-level supervision that is closely tied to the final task objective.

\section{Implementation and Environment Details}

\label{app:implementation_details}

\paragraph{Hardware and software environment.}

All training experiments are conducted on a single node with 8 NVIDIA H800 GPUs. Each GPU has approximately 80GB memory. The NVIDIA driver version is 570.148.08, and the CUDA version reported by \texttt{nvidia-smi} is 12.8. We use \texttt{verl} as the PPO training framework and vLLM as the rollout backend for agentic search generation.

\paragraph{Training setup.}

We train OASES with PPO on the combined HotpotQA and 2WikiMultihopQA training set. Unless otherwise specified, the validation set is selected from the target evaluation benchmark. The training batch size is set to 256, and the validation batch size is also set to 256. We train for 2 epochs. The maximum prompt length is 3000 tokens, and the maximum response length is 2000 tokens.

\paragraph{Model and optimization hyperparameters.}

We use Qwen-series base models as the policy and critic backbone. The actor learning rate is set to $2\times10^{-7}$, and the critic learning rate is set to $1\times10^{-6}$. The PPO mini-batch size is 32, with a PPO micro-batch size of 2 per GPU. We enable gradient checkpointing and remove-padding optimization for both the actor and critic models. We do not use KL loss in the actor objective or KL penalty in the reward.

\paragraph{Rollout and retrieval configuration.}

We use the \texttt{vllm\_agentic} rollout engine with tensor parallel size 1 and GPU memory utilization set to 0.6. The agent uses an external search tool for multi-turn retrieval. In our implementation, the search service is deployed as multiple parallel retrieval endpoints, and the agent queries these endpoints during rollout generation.

\paragraph{Reward configuration.}

Both the final outcome reward and the process reward are computed using F1 against the ground-truth answer. The default process-reward weight is set to $\alpha=0.5$. For ablation experiments on the process-reward weight, we vary $\alpha$ as described in Figure~\ref{fig:alpha_ablation}. 

\paragraph{Evaluation.}

We evaluate model performance using exact match (EM) and token-level F1. During training, validation and checkpointing are performed every fixed number of PPO updates, determined by the training batch size and dataset size.

\section{Efficiency Analysis of State-Evaluation Rollouts}
\label{app:efficiency}

OASES introduces additional state-evaluation rollouts during training. To quantify the resulting overhead, we report the generated-token cost of the main search rollout and the state-evaluation rollouts after convergence. We focus on generated tokens because they directly reflect the decoding cost during rollout collection. As shown in Table~\ref{tab:efficiency}, although each query requires an average of $2.76$ state-evaluation rollouts, each evaluation rollout is short, generating only $66.31$ tokens on average. Consequently, state-evaluation rollouts account for only $14.94\%$ of the generated-token cost per query, while the main search rollout accounts for $85.06\%$. This indicates that the additional evaluation rollouts provide dense training supervision with relatively lightweight decoding overhead.

\begin{table}[t]
\centering
\small
\setlength{\tabcolsep}{6pt}
\renewcommand{\arraystretch}{1.15}
\caption{Generated-token cost of search and state-evaluation rollouts after convergence.}
\label{tab:efficiency}
\begin{tabular}{lcccc}
\toprule
\textbf{Component}
& \textbf{\makecell{Avg. \# Rollouts\\per Query}}
& \textbf{\makecell{Avg. Generated\\Tokens per Rollout}}
& \textbf{\makecell{Avg. Generated\\Tokens per Query}}
& \textbf{Token Share} \\
\midrule
Search rollout
& 1.00
& 1041.76
& 1041.76
& 85.06\% \\
State-evaluation rollouts
& 2.76
& 66.31
& 183.02
& 14.94\% \\
\midrule
Total
& 3.76
& --
& 1224.78
& 100.00\% \\
\bottomrule
\end{tabular}
\end{table}

\section{Prompt Templates}
\label{app:prompt_templates}

Table~\ref{tab:prompt_templates} shows the prompt templates used in OASES. The main search prompt guides the agent to perform multi-step search, while the state-evaluation prompts assess whether each search state supports answering the original question. For state evaluation, we use a question-only prompt for the initial state and a context-aware prompt for later states with accumulated search observations.

\begin{table*}[t]
\centering
\caption{Prompt templates used in OASES, including the main search prompt and two search-state evaluation prompts for question-only and context-aware evaluation.}
\label{tab:prompt_templates}
\small
\begin{tabular}{cc}
\toprule
\textbf{Main Search Prompt} 
& \textbf{Search-state Evaluation Prompts} \\
\midrule
\begin{minipage}[t]{0.6\textwidth}
\vspace{0pt}
\begin{tcolorbox}[
    colback=gray!4,
    colframe=gray!45,
    boxrule=0.5pt,
    arc=2pt,
    left=3pt,
    right=3pt,
    top=3pt,
    bottom=3pt,
    width=\linewidth
]
\scriptsize
\begin{verbatim}
You are an intelligent Research Agent designed to answer questions 
by retrieving information from the internet.

## Operational Logic
1. Plan: Analyze the user's question. If the answer requires 
   multiple steps, break it down.
2. Search: Use the `<search>` tag to query the search engine. Keep 
   queries specific and efficient.
3. Observe: Wait for the `<result>` output provided by the system.
4. Iterate: Analyze the search results in a `<think>` block.
   - If information is missing, formulate a new search query based 
     on what you found.
   - If the information is sufficient, proceed to answer.
5. Answer: Output the final answer in an `<answer>` tag. Keep it 
   concise and direct.

## Response Format
- Thinking Process: <think> ... </think>
- Search Action: <search> ... </search>
- Final Answer: <answer> ... </answer>

## Constraints
- Stop after Search: Do not generate `<result>` tags yourself. 
  Wait for the system.
- Search Limit: There is a maximum limit on the number of 
  search rounds. If the limit is reached, stop searching and 
  output the final answer using available information.
\end{verbatim}
\end{tcolorbox}
\end{minipage}
&
\begin{minipage}[t]{0.35\textwidth}
\vspace{0pt}

\begin{tcolorbox}[
    title=\textbf{Question-Only Evaluation},
    colback=gray!4,
    colframe=gray!45,
    boxrule=0.5pt,
    arc=2pt,
    left=3pt,
    right=3pt,
    top=3pt,
    bottom=3pt,
    width=\linewidth
]
\scriptsize
\begin{verbatim}
You are an expert AI assistant 
capable of answering questions 
precisely. 
Answer the user's question based 
on your internal knowledge.

You must provide a concise answer 
enclosed within specific tags as 
follows:
<eval> The Answer </eval>
\end{verbatim}
\end{tcolorbox}

\vfill

\begin{tcolorbox}[
    title=\textbf{Context-Aware Evaluation},
    colback=gray!4,
    colframe=gray!45,
    boxrule=0.5pt,
    arc=2pt,
    left=3pt,
    right=3pt,
    top=3pt,
    bottom=3pt,
    width=\linewidth
]
\scriptsize
\begin{verbatim}
You are an expert AI assistant 
capable of answering questions 
precisely. 
Answer the user's question 
using the provided Context and 
your internal knowledge.

You must provide a concise answer 
enclosed within specific tags as 
follows:
<eval> The Answer </eval>
\end{verbatim}
\end{tcolorbox}

\end{minipage}
\vspace{5pt}
\\
\bottomrule
\end{tabular}
\end{table*}

\section{Baseline Details}

\label{app:baseline_details}

This section provides additional details about the baselines used in our experiments. 

\paragraph{Direct LLM.}

Direct LLM is a non-retrieval baseline where the model directly answers the question using only its parametric knowledge. It does not access external documents or invoke search tools. 

\paragraph{Naive RAG.}

Naive RAG retrieves relevant passages from the corpus and appends them to the model input before answer generation. Unlike agentic search methods, it follows a static retrieve-then-read pipeline and does not allow the model to iteratively refine queries, inspect intermediate evidence, or decide whether additional search is needed~\cite{lewis2020retrieval}. 

\paragraph{Search-o1.}

Search-o1 is an agentic search-enhanced reasoning framework that augments large reasoning models with dynamic retrieval during the reasoning process. It introduces an agentic RAG workflow that allows the model to retrieve external knowledge when encountering uncertain knowledge points, and uses a Reason-in-Documents module to analyze retrieved documents before injecting information back into the reasoning chain~\cite{li2025search}. Search-o1 is included as a representative inference-time agentic search baseline.

\paragraph{Search-R1.}

Search-R1 trains language models to interleave reasoning and search through reinforcement learning. The model learns to autonomously generate search queries during step-by-step reasoning and interact with a real-time retrieval system. Its training relies on outcome-based rewards and retrieved-token masking for stable RL optimization~\cite{jin2025search}.

\paragraph{R1-Searcher.}

R1-Searcher is a two-stage outcome-based RL framework designed to incentivize search capability in LLMs. It first teaches the model when and how to invoke external search, and then optimizes the model to use retrieved information for final answer generation~\cite{song2025r1}. 

\paragraph{StepSearch.}

StepSearch trains search-augmented LLMs with step-wise PPO and adds search-step rewards to the final answer reward~\cite{zheng2025stepsearch}. GPT-4o is used to construct subquestion--answer pairs and search queries, which are then treated as gold supervision for reward computation. Based on this supervision, the process reward is derived from retrieval-side signals: each search step is rewarded by the marginal improvement in retrieved-document coverage over gold information documents, measured with TF--IDF cosine similarity, and penalized for retrieving redundant documents.

\paragraph{ReasonRAG.}

ReasonRAG studies process-supervised training for agentic RAG~\cite{zhang2025process}. Instead of relying only on final-answer correctness, it constructs process-supervised rollouts with Monte Carlo Tree Search (MCTS), obtaining preference signals over intermediate RAG actions such as query generation, evidence extraction, and answer generation. These process-level preferences are then optimized with DPO to teach the model better search and evidence-use strategies. 

\paragraph{TIPS.}

TIPS proposes turn-level information-potential reward shaping for search-augmented LLMs~\cite{xie2026tips}. Its reward is computed from the increase in the teacher model's likelihood of the correct answer after each reasoning--tool-call turn. Specifically, TIPS treats this likelihood as an information potential and assigns a turn-level shaping reward according to how much the current turn increases that potential. 

\paragraph{SubSearch.}

SubSearch trains search agents with intrinsic intermediate rewards for retrieval tasks~\cite{petcu2026subsearch}. Its process rewards come from two internally computed signals. First, a subquery answerability reward measures how well the top-$k$ retrieved documents address each generated subquery, using embedding cosine similarity between the subquery and retrieved documents. Second, a decomposition reward evaluates whether the generated subqueries preserve the semantic coverage of the parent query while remaining distinct from each other. These intermediate rewards are combined with the final exact-match answer reward and a format reward.

\section{Limitations}

\label{app:limitations}

\paragraph{Increased training cost.} 

OASES introduces additional computational overhead during training. Besides generating the main search trajectory, it performs state-conditioned evaluation rollouts for intermediate search states. As a result, rollout collection becomes slower and the total training time increases compared with methods that rely only on sparse outcome rewards.

\paragraph{Limited applicability to search scenarios.} 

OASES is currently designed for agentic search tasks, where intermediate states can be naturally defined by the accumulated retrieved evidence. Therefore, it may not directly apply to tasks without external search, clearly identifiable intermediate states, or a meaningful notion of evidence accumulation.

\paragraph{Dependence on outcome evaluation.} 

The process supervision in OASES is derived from outcome improvements over search states. Thus, its effectiveness depends on the reliability of the final outcome evaluator. If the outcome reward is noisy, biased, or incomplete, the induced intermediate rewards may also be affected.

\begin{table*}[t]
\centering
\scriptsize
\setlength{\tabcolsep}{3pt}
\renewcommand{\arraystretch}{1.05}
\caption{Negative case study showing degraded search-state evaluation after search-only training.}
\label{tab:bad_case_search_only}

\begin{tabularx}{\linewidth}{p{0.06\linewidth}|p{0.12\linewidth}|X}
\toprule
\multicolumn{3}{p{\linewidth}}{
\textbf{Query:} What was the first use of the East Sitting Hall at the official residence and workplace of the President of the United States?
} \\
\multicolumn{3}{p{\linewidth}}{
\textbf{Ground Truth:} reception room
} \\
\midrule
\textbf{Turn} & \textbf{Component} & \textbf{Output / Search-State Evaluation} \\
\midrule

Turn 0
& \cellcolor{gray!12} Eval.
& \cellcolor{gray!12}
\textbf{Eval:} \texttt{<eval> Congress held its first full session </eval>} \newline
\textbf{Score:} $\gamma_0=0.000$
\quad
\textcolor{red}{\textbf{(\ding{55} unsupported)}}
\\
\midrule

\multirow{3}{*}{Turn 1}
& Search
& \texttt{<search> official residence and workplace of the President of the United States </search>}
\\
\cmidrule(lr){2-3}
& Result
& The official residence and workplace of the U.S. President is the \textbf{White House}.
\\
\cmidrule(lr){2-3}
& \cellcolor{gray!12} Eval.
& \cellcolor{gray!12}
\textbf{Eval:} \texttt{<eval> 1792 </eval>} \newline
\textbf{Score:} $\gamma_1=0.000$
\quad
\textcolor{red}{\textbf{(\ding{55} distracted by irrelevant evidence)}} \newline
\textbf{Source:} \textit{``Construction took place between 1792 and 1800 using Aquia...''}
\\
\midrule

\multirow{3}{*}{Turn 2}
& Search
& \texttt{<search> first use of East Sitting Hall in White House </search>}
\\
\cmidrule(lr){2-3}
& Result
& The East Sitting Hall was first used as a \textbf{reception room for guests of the president}.
\\
\cmidrule(lr){2-3}
& \cellcolor{gray!12} Eval.
& \cellcolor{gray!12}
\textbf{Eval:} \texttt{<eval> 2,161,000 </eval>} \newline
\textbf{Score:} $\gamma_2=0.000$
\quad
\textcolor{red}{\textbf{(\ding{55} copied from prompt example)}} \newline
\textbf{Source:} Prompt example: \textit{``The population is approximately 2,161,000. Now I will...''}
\\
\midrule

Turn 3
& \cellcolor{gray!12} Answer
& \cellcolor{gray!12}
\textbf{Answer:} \texttt{<answer> reception room</answer>} \newline
\textbf{Outcome:} $\gamma_{\mathrm{out}}=1.000$
\quad
\textcolor{green!50!black}{\textbf{(\checkmark\ correct)}}
\\
\midrule

\multicolumn{3}{p{\linewidth}}{
\textbf{Failure Pattern:} The model obtains a correct final answer, but all state-conditioned evaluations are incorrect.
} \\
\bottomrule
\end{tabularx}
\end{table*}

\begin{table*}[t]
\centering
\small
\setlength{\tabcolsep}{5pt}
\renewcommand{\arraystretch}{1.12}
\caption{Positive case study of outcome-aligned process reward in OASES.}
\label{tab:case_study}
\resizebox{\linewidth}{!}{
\begin{tabular}{p{0.08\linewidth}|p{0.13\linewidth}|p{0.73\linewidth}}
\toprule
\multicolumn{3}{l}{\textbf{Query:} When is the director of film \textit{The Opposite of Sex}'s birthday?} \\
\multicolumn{3}{l}{\textbf{Ground Truth:} April 14, 1955} \\
\midrule
\textbf{Turn} & \textbf{Component} & \textbf{Output / Search-State Evaluation} \\
\midrule

Turn 0
& \cellcolor{gray!12}Evaluation of $s_0$
& \cellcolor{gray!12}
\begin{tabular}[t]{@{}l@{}}
\textbf{Eval Answer:} \texttt{<eval> April 20, 1962 </eval>} \\
\textbf{Score:} $\gamma_0=0.333$
\end{tabular}
\\
\midrule

\multirow{4}{*}{Turn 1}
& Think
& \texttt{<thinking> I need to answer this in two steps. First, I must identify the director of the film ``The Opposite of Sex''. Second, I need to find the birth date of that specific director. </thinking>} \\
\cmidrule(lr){2-3}
& Search
& \texttt{<search> director of film The Opposite of Sex </search>} \\
\cmidrule(lr){2-3}
& Result
& \texttt{<result>} The Opposite of Sex is a 1998 American romantic comedy film 
\textcolor{red}{written and directed by Don Roos} and starring Christina Ricci, Martin Donovan and Lisa Kudrow.
Don Roos is well known for his work as a screenwriter... \texttt{</result>} \\
\cmidrule(lr){2-3}
& \cellcolor{gray!12}Evaluation of $s_1$
& \cellcolor{gray!12}
\begin{tabular}[t]{@{}l@{}}
\textbf{Eval Answer:} \texttt{<eval> April 14, 1950 </eval>} \\
\textbf{Score:} $\gamma_1=0.667$ \quad \textbf{Search-State Gain:} $\gamma_1-\gamma_0=+0.334$
\end{tabular}
\\

\midrule

\multirow{4}{*}{Turn 2}
& Think
& \texttt{<thinking> The director of the film ``The Opposite of Sex'' is Don Roos. Now I need to find his birth date. </thinking>} \\
\cmidrule(lr){2-3}
& Search
& \texttt{<search> Don Roos birth date </search>} \\
\cmidrule(lr){2-3}
& Result
& \texttt{<result>} Don Roos Donald Paul ``Don'' Roos 
\textcolor{red}{(born April 14, 1955)} is an American screenwriter and film director. Roos was born in New York. He attended the University of Notre Dame in Indiana and... \texttt{</result>} \\
\cmidrule(lr){2-3}
& \cellcolor{gray!12}Evaluation of $s_2$
& \cellcolor{gray!12}
\begin{tabular}[t]{@{}l@{}}
\textbf{Eval Answer:} \texttt{<eval> April 14, 1955 </eval>} \\
\textbf{Score:} $\gamma_2=1.000$ \quad \textbf{Search-State Gain:} $\gamma_2-\gamma_1=+0.333$
\end{tabular}
\\
\midrule

\multirow{2}{*}{Turn 3}
& Think
& \texttt{<thinking> The search results confirm that Don Roos was born on April 14, 1955. I have the final answer. </thinking>} \\
\cmidrule(lr){2-3}
& \cellcolor{gray!12}Answer

& \cellcolor{gray!12}
\begin{tabular}[t]{@{}l@{}}
\texttt{<answer> April 14, 1955 </answer>} \\
\textbf{Outcome Reward:} $r^{\mathrm{out}}=\gamma_{\mathrm{out}}=1.000$
\end{tabular}
\\
\midrule
\multicolumn{3}{l}{
\textbf{Final Prediction:} \texttt{April 14, 1955}
\quad
\textcolor{green!50!black}{\textbf{(\checkmark\ Matches Ground Truth)}}
} \\
\bottomrule
\end{tabular}
}
\end{table*}

\begin{table*}[t]
\centering
\small
\setlength{\tabcolsep}{5pt}
\renewcommand{\arraystretch}{1.12}
\caption{Negative case study of outcome-aligned process reward in OASES.}
\label{tab:negative_case_study}
\resizebox{\linewidth}{!}{
\begin{tabular}{p{0.08\linewidth}|p{0.13\linewidth}|p{0.73\linewidth}}
\toprule
\multicolumn{3}{l}{\textbf{Query:} Where did the director of film \textit{Legally Blonde} study?} \\
\multicolumn{3}{l}{\textbf{Ground Truth:} California Institute of the Arts} \\
\midrule
\textbf{Turn} & \textbf{Component} & \textbf{Output / Search-State Evaluation} \\
\midrule

Turn 0
& \cellcolor{gray!12}Evaluation of $s_0$
& \cellcolor{gray!12}
\begin{tabular}[t]{@{}l@{}}
\textbf{Eval Answer:} \texttt{<eval> University of Southern California (USC) </eval>} \\
\textbf{Score:} $\gamma_0=0.444$
\end{tabular}
\\
\midrule

\multirow{4}{*}{Turn 1}
& Think
& \texttt{<thinking> I need to answer this in two steps. First, I must identify the director of the film ``Legally Blonde''. Second, I need to find the educational background of that specific director. </thinking>} \\
\cmidrule(lr){2-3}
& Search
& \texttt{<search> director of film Legally Blondes </search>} \\
\cmidrule(lr){2-3}
& Result
& \texttt{<result>} Legally Blonde is a 2001 American comedy film based on Amanda Brown's novel of the same name. It was 
\textcolor{red}{directed by Robert Luketic}, scripted by Karen McCullah Lutz and Kirsten Smith, and stars Reese Witherspoon, Luke Wilson, Selma Blair, Matthew Davis, Victor Garber... \texttt{</result>} \\
\cmidrule(lr){2-3}
& \cellcolor{gray!12}Evaluation of $s_1$
& \cellcolor{gray!12}
\begin{tabular}[t]{@{}l@{}}
\textbf{Eval Answer:} \texttt{<eval> University of Melbourne </eval>} \\
\textbf{Score:} $\gamma_1=0.286$ \quad \textbf{Search-State Gain:} $\gamma_1-\gamma_0=-0.158$
\end{tabular}
\\

\midrule

\multirow{4}{*}{Turn 2}
& Think
& \texttt{<thinking> The director of the film ``Legally Blonde'' is Robert Luketic. Now I need to find out where he studied. </thinking>} \\
\cmidrule(lr){2-3}
& Search
& \texttt{<search> Robert Luketic education </search>} \\
\cmidrule(lr){2-3}
& Result
& \texttt{<result>} The retrieved results do not match Robert Luketic. They instead mention unrelated entities such as Robert Lukens and Robert S. Lasnik. One snippet states that 
\textcolor{red}{Robert Lukens received his bachelor's degree from Temple University}, while another describes Robert S. Lasnik's education. \texttt{</result>} \\
\cmidrule(lr){2-3}
& \cellcolor{gray!12}Evaluation of $s_2$
& \cellcolor{gray!12}
\begin{tabular}[t]{@{}l@{}}
\textbf{Eval Answer:} \texttt{<eval> Temple University </eval>} \\
\textbf{Score:} $\gamma_2=0.000$ \quad \textbf{Search-State Gain:} $\gamma_2-\gamma_1=-0.286$
\end{tabular}
\\
\midrule

\multirow{2}{*}{Turn 3}
& Think
& \texttt{<thinking> The search results indicate that Robert Luketic studied at the University of California, Los Angeles (UCLA). </thinking>} \\
\cmidrule(lr){2-3}
& \cellcolor{gray!12}Answer
& \cellcolor{gray!12}
\begin{tabular}[t]{@{}l@{}}
\texttt{<answer> UCLA </answer>} \\
\textbf{Outcome Reward:} $r^{\mathrm{out}}=\gamma_{\mathrm{out}}=0.000$
\end{tabular}
\\
\midrule
\multicolumn{3}{l}{
\textbf{Final Prediction:} \texttt{UCLA}
\quad
\textcolor{red}{\textbf{(\ding{55} Does Not Match Ground Truth)}}
} \\
\bottomrule
\end{tabular}
}
\end{table*}

\section{Failure Case of Search-Only Training}
\label{app:search_only_case}

Table~\ref{tab:bad_case_search_only} shows a failure case after search-only training.
Although the model eventually produces the correct final answer, its state-conditioned evaluations are incorrect at every intermediate state.
At Turn~0, the evaluator relies on unsupported internal knowledge; after the first search, it is distracted by irrelevant construction-year evidence; after the second search, it copies an unrelated answer from the prompt example despite having retrieved the correct evidence.
This pattern suggests that optimizing only the search trajectory can improve final-answer behavior while leaving state evaluation unreliable.
Such failures motivate search--evaluation co-training in OASES, which jointly trains the policy to search and to evaluate whether its accumulated information supports the final answer.

\section{Discussion: Why Joint Search--Evaluation Optimization?}
\label{sec:joint_search_eval_discussion}

Likelihood-based information-gain methods turn intermediate evaluation into a direct conditional-answering task, which differs from multi-turn agentic search~\cite{wang2025information, liang2026ig, hu2026optimizing}. In agentic search, the model decides whether to continue searching or stop and answer, and the final answer is generated after the policy's own reasoning and stopping decision. By contrast, likelihood-based scoring asks the model to score the golden answer from a fixed intermediate state, often with a manually appended prefix such as ``Now there is enough information to answer''~\cite{wang2025information}. Such prefixes inject an artificial stopping decision, while omitting them creates a context that differs from the model's usual ReAct-style answer-generation format. 

More importantly, likelihood-based scoring is difficult to jointly optimize as an explicit evaluation behavior. Since the score is computed from the likelihood of the gold answer, directly training the evaluator to increase this likelihood under intermediate search states would expose the evaluator to the target answer before sufficient evidence is available. This creates a shortcut: the evaluator can improve the scoring objective by memorizing or fitting the gold answer, rather than learning to judge whether the accumulated evidence actually supports answering the question. As a result, the optimized evaluator may assign high scores to under-informed states, weakening the intended role of information gain as a measure of evidence usefulness. In addition, likelihood scoring provides only a scalar probability of a known answer, rather than a generated state-conditioned answer whose quality can be verified by the outcome metric. Therefore, it does not naturally produce trainable evaluation rollouts that match the policy's answer-generation behavior.

As RL training progresses, this mismatch can grow because the policy may specialize in answering only after its own search and reasoning trajectory, whereas likelihood-based evaluation continues to score gold-answer probability under artificial or partial contexts. Consequently, the estimated likelihood gain may not accurately reflect whether the current state helps the policy produce a better final answer.

OASES instead evaluates each intermediate state by generating a state-conditioned answer with the same policy and scoring it using the verifiable outcome signal (e.g. the F1 or EM scores). This turns intermediate evaluation into an explicit generation behavior rather than a likelihood-scoring heuristic. Given the evidence accumulated so far, the policy directly tests whether the current state can support a correct answer, and the resulting score is computed in the same output space as the final task objective. Therefore, the induced process reward measures outcome improvement across search states, rather than a proxy change in answer likelihood.

More importantly, OASES jointly optimizes search and state evaluation with shared parameters. This allows the model to learn two complementary abilities at the same time: acquiring useful evidence through multi-turn search and judging whether the evidence collected so far is sufficient for answering. As the search policy evolves during RL, the evaluator evolves with it, keeping intermediate state scores compatible with the policy's current generation behavior. As a result, OASES provides process rewards that are both outcome-aligned and policy-adaptive, leading to more reliable credit assignment for intermediate search decisions. Without such joint optimization, search-only RL may make the policy increasingly specialized for final-answer generation after complete search trajectories, while weakening its ability to answer from partial states; this evaluator forgetting can make state scores unreliable and cause process-reward supervision to fail.

\section{Case Studies of Outcome-Aligned Process Rewards}
\label{app:case_studies}

We provide two case studies to illustrate how OASES assigns stepwise process rewards from outcome-aligned improvements across search states.
For clarity, Tables~\ref{tab:case_study} and~\ref{tab:negative_case_study} report the unscaled score gain $\gamma_t-\gamma_{t-1}$ at each search turn. 
During training, this gain is multiplied by the process-reward weight $\alpha$ to obtain the actual process reward $r_t^{\mathrm{proc}}=\alpha(\gamma_t-\gamma_{t-1})$.

\paragraph{Positive case.}
Table~\ref{tab:case_study} shows a successful two-hop search trajectory. 
The query asks for the birthday of the director of \textit{The Opposite of Sex}. 
At the initial state $s_0$, the evaluator only sees the question and predicts an incorrect date, yielding $\gamma_0=0.333$. 
After the first search, the trajectory identifies the bridge entity Don Roos as the director of the film. 
Although the target birth date has not yet been retrieved, this search state becomes more informative, and the evaluator answer improves from \texttt{April 20, 1962} to \texttt{April 14, 1950}, increasing the score to $\gamma_1=0.667$. 
This produces a positive search-state gain of $\gamma_1-\gamma_0=+0.334$, rewarding the first search turn for finding the key entity.

The second search directly retrieves Don Roos's birth date. 
With this evidence available in $s_2$, the evaluator outputs the exact answer \texttt{April 14, 1955}, and the score reaches $\gamma_2=1.000$. 
This yields another positive search-state gain of $\gamma_2-\gamma_1=+0.333$. 
The final answer also matches the ground truth, so the main trajectory receives the outcome reward $r^{\mathrm{out}}=\gamma_{\mathrm{out}}=1.000$. 
This case illustrates the intended behavior of OASES: search turns are rewarded according to how much they improve the answerability of the current search state. 
The first turn discovers the bridge entity, while the second turn retrieves the target attribute.

\paragraph{Negative case.}
Table~\ref{tab:negative_case_study} shows a contrasting case where search introduces misleading evidence. 
The query asks where the director of \textit{Legally Blonde} studied, whose ground-truth answer is \textit{California Institute of the Arts}. 
At the initial state $s_0$, the evaluator predicts \texttt{University of Southern California (USC)}, obtaining $\gamma_0=0.444$ due to partial token overlap with the ground truth. 
The first search correctly identifies Robert Luketic as the director, but the evaluator still predicts an incorrect institution, \texttt{University of Melbourne}. 
The score drops to $\gamma_1=0.286$, giving a negative search-state gain of $\gamma_1-\gamma_0=-0.158$.

The second search is more harmful: instead of retrieving Robert Luketic's education, it returns snippets about unrelated entities such as Robert Lukens and Robert S. Lasnik. 
The evaluator is then misled to answer \texttt{Temple University}, and the score decreases further to $\gamma_2=0.000$. 
This produces another negative search-state gain of $\gamma_2-\gamma_1=-0.286$. 
The final model answer is \texttt{UCLA}, which does not match the ground truth, so the trajectory receives no outcome reward. 
This case shows that OASES does not blindly reward additional search. 
When a search action makes the intermediate state less answerable by introducing irrelevant or misleading evidence, the corresponding score gain becomes negative, resulting in negative process feedback for that search turn.

\paragraph{Summary.}
Together, the two cases demonstrate that OASES provides outcome-aligned credit assignment for agentic search. Search steps that uncover bridge entities or target attributes receive positive process rewards when they improve state answerability, while misleading steps receive negative feedback when they degrade it. This dense feedback complements the sparse final outcome reward and helps the policy learn which search actions are useful for solving the original question.

\newpage

\newpage
\section*{NeurIPS Paper Checklist}

\begin{enumerate}

\item {\bf Claims}
    \item[] Question: Do the main claims made in the abstract and introduction accurately reflect the paper's contributions and scope?
    \item[] Answer: \answerYes{} 
    \item[] Justification: : We claimed the main contribution and the scope in our abstract and introduction.

\item {\bf Limitations}
    \item[] Question: Does the paper discuss the limitations of the work performed by the authors?
    \item[] Answer: \answerYes{} 
    \item[] Justification: We discuss some limitations of our proposed method in Appendix~\ref{app:limitations}.

\item {\bf Theory assumptions and proofs}
    \item[] Question: For each theoretical result, does the paper provide the full set of assumptions and a complete (and correct) proof?
    \item[] Answer: \answerNA{} 
    \item[] Justification: The paper does not include theoretical results

    \item {\bf Experimental result reproducibility}
    \item[] Question: Does the paper fully disclose all the information needed to reproduce the main experimental results of the paper to the extent that it affects the main claims and/or conclusions of the paper (regardless of whether the code and data are provided or not)?
    \item[] Answer: \answerYes{} 
    \item[] Justification: We provide all the details and the anonymous code url in the paper.

\item {\bf Open access to data and code}
    \item[] Question: Does the paper provide open access to the data and code, with sufficient instructions to faithfully reproduce the main experimental results, as described in supplemental material?
    \item[] Answer: \answerYes{} 
    \item[] Justification: The paper provides an anonymous code release link, and the released repository includes the implementation and instructions needed to reproduce the main results.

\item {\bf Experimental setting/details}
    \item[] Question: Does the paper specify all the training and test details (e.g., data splits, hyperparameters, how they were chosen, type of optimizer) necessary to understand the results?
    \item[] Answer: \answerYes{} 
    \item[] Justification:  We provide many details of the experiments and the key experimental settings.

\item {\bf Experiment statistical significance}
    \item[] Question: Does the paper report error bars suitably and correctly defined or other appropriate information about the statistical significance of the experiments?
    \item[] Answer: \answerNo{} 
    \item[] Justification: We don’t do the statistical significance tests.

\item {\bf Experiments compute resources}
    \item[] Question: For each experiment, does the paper provide sufficient information on the computer resources (type of compute workers, memory, time of execution) needed to reproduce the experiments?
    \item[] Answer: \answerYes{} 
    \item[] Justification: We discuss experiments compute resources in Appendix~\ref{app:implementation_details}.
    
\item {\bf Code of ethics}
    \item[] Question: Does the research conducted in the paper conform, in every respect, with the NeurIPS Code of Ethics \url{https://neurips.cc/public/EthicsGuidelines}?
    \item[] Answer: \answerYes{} 
    \item[] Justification: No ethics problem.

\item {\bf Broader impacts}
    \item[] Question: Does the paper discuss both potential positive societal impacts and negative societal impacts of the work performed?
    \item[] Answer: \answerNA{} 
    \item[] Justification: This discussion is not necessary for our work.
    
\item {\bf Safeguards}
    \item[] Question: Does the paper describe safeguards that have been put in place for responsible release of data or models that have a high risk for misuse (e.g., pre-trained language models, image generators, or scraped datasets)?
    \item[] Answer: \answerNA{} 
    \item[] Justification:  Our work poses no such risks.

\item {\bf Licenses for existing assets}
    \item[] Question: Are the creators or original owners of assets (e.g., code, data, models), used in the paper, properly credited and are the license and terms of use explicitly mentioned and properly respected?
    \item[] Answer: \answerYes{} 
    \item[] Justification:  Our paper meets this requirement.

\item {\bf New assets}
    \item[] Question: Are new assets introduced in the paper well documented and is the documentation provided alongside the assets?
    \item[] Answer: \answerNA{} 
    \item[] Justification: We don’t release new assets.

\item {\bf Crowdsourcing and research with human subjects}
    \item[] Question: For crowdsourcing experiments and research with human subjects, does the paper include the full text of instructions given to participants and screenshots, if applicable, as well as details about compensation (if any)? 
    \item[] Answer: \answerNA{} 
    \item[] Justification:  Our paper does not involve crowdsourcing nor research with human subjects.

\item {\bf Institutional review board (IRB) approvals or equivalent for research with human subjects}
    \item[] Question: Does the paper describe potential risks incurred by study participants, whether such risks were disclosed to the subjects, and whether Institutional Review Board (IRB) approvals (or an equivalent approval/review based on the requirements of your country or institution) were obtained?
    \item[] Answer: \answerNA{} 
    \item[] Justification: Our paper does not involve crowdsourcing nor research with human subjects.

\item {\bf Declaration of LLM usage}
    \item[] Question: Does the paper describe the usage of LLMs if it is an important, original, or non-standard component of the core methods in this research? Note that if the LLM is used only for writing, editing, or formatting purposes and does \emph{not} impact the core methodology, scientific rigor, or originality of the research, declaration is not required.
    \item[] Answer: \answerNA{} 
    \item[] Justification:  We only checked for syntax errors using LLM.

\end{enumerate}

\end{document}